\def\year{2022}\relax
\documentclass[letterpaper]{article} % DO NOT CHANGE THIS
\usepackage{aaai22}  % DO NOT CHANGE THIS
\usepackage{times}  % DO NOT CHANGE THIS
\usepackage{helvet}  % DO NOT CHANGE THIS
\usepackage{courier}  % DO NOT CHANGE THIS
\usepackage[hyphens]{url}  % DO NOT CHANGE THIS
\usepackage{graphicx} % DO NOT CHANGE THIS
\usepackage{natbib}  % DO NOT CHANGE THIS AND DO NOT ADD ANY OPTIONS TO IT
\usepackage{caption} % DO NOT CHANGE THIS AND DO NOT ADD ANY OPTIONS TO IT
\DeclareCaptionStyle{ruled}{labelfont=normalfont,labelsep=colon,strut=off} % DO NOT CHANGE THIS
\usepackage{algorithm}
\usepackage{algorithmic}
\usepackage{bbm}
\usepackage{amssymb}
\usepackage{mathtools}

\DeclareMathOperator*{\argmax}{arg\,max} % * allows typesetting beneath

\usepackage{booktabs}
\usepackage{dsfont}

\newcommand{\STAB}[1]{\begin{tabular}{@{}c@{}}#1\end{tabular}}
\usepackage{multirow}

\usepackage{newfloat}
\usepackage{listings}
\floatstyle{ruled}
\newfloat{listing}{tb}{lst}{}
\floatname{listing}{Listing}
\newcommand*\samethanks[1][\value{footnote}]{\footnotemark[#1]}

\title{Eliciting and Learning with Soft Labels from Every Annotator}
\author {
    Katherine M. Collins\thanks{Both authors contributed equally.}\textsuperscript{\rm 1},
    Umang Bhatt\samethanks\textsuperscript{\rm 1, 2},
    Adrian Weller\textsuperscript{\rm 1,2}
}
\affiliations {
    \textsuperscript{\rm 1} University of Cambridge\\
    \textsuperscript{\rm 2} The Alan Turing Institute\\
    \{kmc61, usb20, aw665\}@cam.ac.uk
}

\begin{document}

\maketitle

\begin{abstract}
The labels used to train machine learning (ML) models are of paramount importance. Typically for ML classification tasks, datasets contain hard labels, yet learning using soft labels has been shown to yield benefits for model generalization, robustness, and calibration.
Earlier work found success in forming soft labels from multiple annotators' hard labels; however, this approach may not converge to the best labels and necessitates many annotators, which can be expensive and inefficient. We focus on efficiently eliciting soft labels from individual annotators. 
We collect and release a dataset of soft labels (which we call \texttt{CIFAR-10S}) over the \texttt{CIFAR-10} test set via a crowdsourcing study ($N=248$). We demonstrate that learning with our labels achieves comparable model performance to prior approaches while requiring far fewer annotators -- albeit with significant temporal costs per elicitation. Our elicitation methodology therefore shows nuanced promise in enabling practitioners to enjoy the benefits of improved model performance and reliability with fewer annotators, and serves as a guide for future dataset curators on the benefits of leveraging richer information, such as categorical uncertainty, from individual annotators.

%The structure and quality of labels used to train machine learning (ML) models are of paramount importance. Typically for ML classification tasks, datasets only contain hard labels, yet learning using soft labels has conferred reams of benefits for model generalization, robustness, and calibration. Unfortunately, there is a little work on constructing information-rich soft labels. Existing work has found success in forming soft labels from multiple annotators' hard labels; however, this approach necessitates many annotators, which can be expensive and inefficient. We focus on efficiently eliciting soft labels from individual annotators by querying annotators' instantaneous uncertainty over inputs. We collect soft labels in a crowdsourcing study ($N=240$) and demonstrate that learning with our labels achieves comparable model performance to prior approaches while requiring far fewer annotators. Our elicitation methodology therefore shows promise towards enabling practitioners to enjoy the benefits of improved model performance and reliability with fewer annotators, and serves as a guide for future dataset curators on the benefits of leveraging richer information, such as categorical uncertainty, from individual annotators.
\end{abstract}

\section{Introduction}

Supervised machine learning (ML) relies on labeled training data.
Most ML datasets for classification are constructed by asking one annotator to provide a single label for an image. However, an annotator might usefully ascribe probabilities to various labels. Requesting %a single judgment by providing a 
just one hard label %for an observation, may require annotators to pick a discrete label from a set of probabilistic judgments. As a result, data labelling might
may be a \textit{lossy} operation, as potentially important information about %the uncertainty in 
an annotator's uncertainty %label 
is not captured. 

\citet{peterson2019human} and \citet{battleday2020capturing} ask multiple annotators each to provide one hard label for every image in the \texttt{CIFAR-10} test set, yielding a label set they call \texttt{CIFAR-10H}.
Soft labels are then obtained by simply aggregating the hard labels over  annotators.
%with annotator-level information rarely released~\cite{prabhakaran2021releasing}.
This set of soft labels is costly to procure as many annotators are required. In addition, while this method indeed captures some notion of probability judgments through multiple annotator labels for a single image, we argue these labels could be misleading since they do not amalgamate individual annotators' soft labels because only the mode judgments from each annotator are aggregated. 

%More meaningful annotation methods which elicit probabilistic judgments \textit{directly} from every annotator offer the potential for annotator-efficient methods which capture individual-level uncertainty over observations and carry the possibility of converging to ``better'' training labels.
% This strat such many-annotator elicitation can be quite costly.

% this is katie on phone - i still have a minute of wifi so looking at this - new changes look awesome!!! thanks :) 

% \citet{davani2022dealing} asked each annotator for a set of plausible labels and assigned probability mass uniformly, but we suggest this is still problematic. %; however, it is unclear how to distribute probability mass over the selected categories. 
% Consider an image which is most likely a dog, possibly a cat, and less likely a deer. An annotator may select all three classes, but it would be perceptually inconsistent to assign a third of the probability mass to all three labels.

\begin{figure*}[htb]
  \begin{center}
  \includegraphics[width=1.0\linewidth]{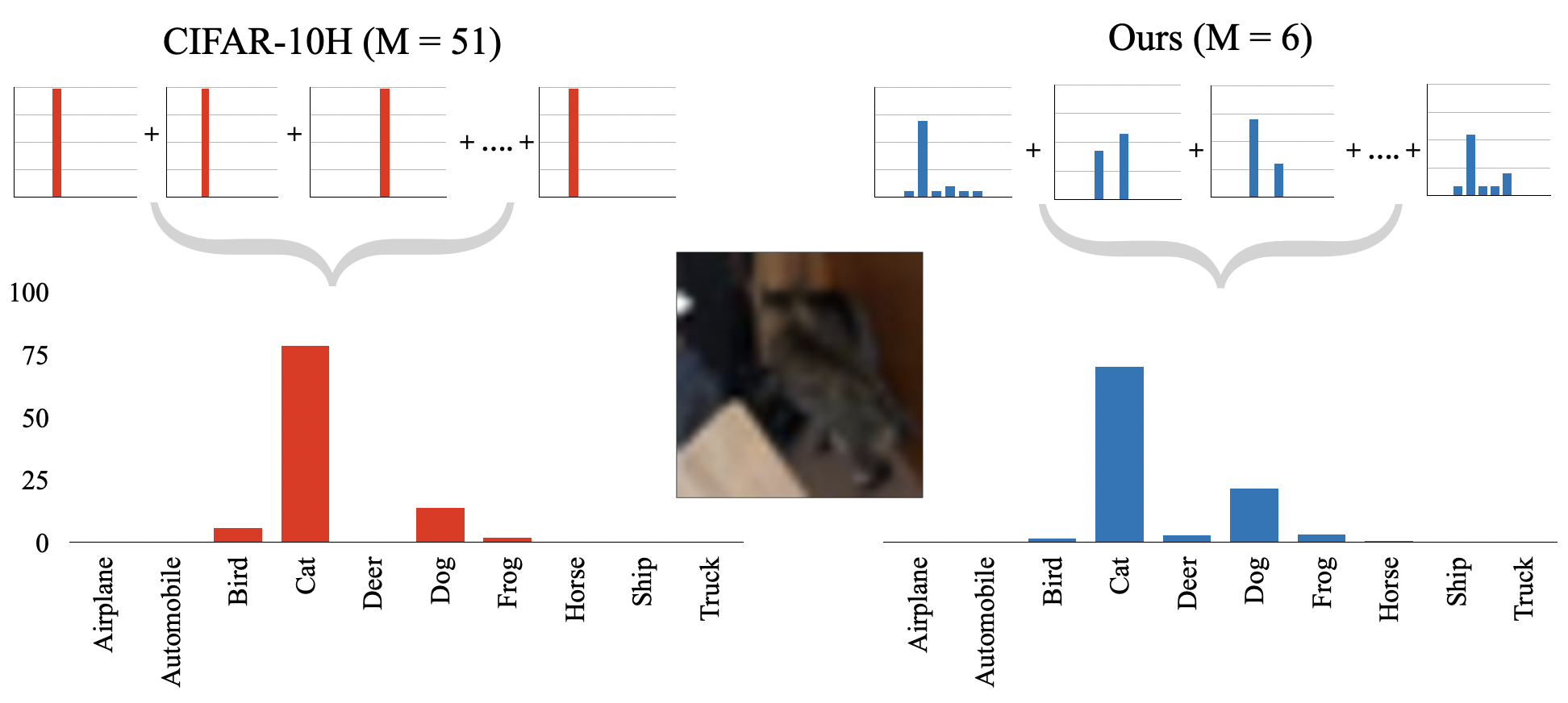}
      \caption{Unlike \texttt{CIFAR-10H}, we enable each annotator to express their uncertainty over image identity, enabling us to converge to richer labels faster: $M = 51$ annotators for~\cite{peterson2019human} vs. $M = 6$ annotators for ours.}
        \label{fig:overview}
  \end{center}
  \end{figure*}

Instead, we %build on \cite{peterson2019human} to 
elicit and aggregate \emph{per-annotator} probabilistic judgments over the label space in an image classification setting, specifically \texttt{CIFAR-10}~\cite{Krizhevsky09learningmultiple}.  Fig.~\ref{fig:overview} illustrates how our method compares to that of \citet{peterson2019human}. 
%We demonstrate that eliciting individual-annotator uncertainty supports more annotator-efficient labeling schemes, while improving the robustness of classifiers trained on the resulting soft labels. 
We highlight the following contributions:
\begin{itemize}
    \item We introduce an efficient approach to elicit soft labels from individual annotators and will release the code for our elicitation interface.
    \item We release our dataset of 6,200 soft labels over 1,000 image datapoints from \texttt{CIFAR-10}. We call this new dataset of \textit{soft} labels \texttt{CIFAR-10S}. 
    \item We show that models trained with \texttt{CIFAR-10S} obtain similar performance (in terms of accuracy, robustness, and calibration) to models trained on~\texttt{CIFAR-10H} with approximately 8.5x fewer annotators.
    %and 60\% of the total annotation time.
\end{itemize}
% % an elicitation strategy for capturing soft labels from individual annotators, and a dataset of 6,050 soft labels over 1,000 datapoints in the test set of \texttt{CIFAR-10}: see Fig. 1. 
% We then show that models trained with our soft labels obtain similar performance (in terms of accuracy, robustness, and calibration) to~\cite{peterson2019human} with approximately \textbf{8.5x fewer annotators} and an estimated \textbf{60\% of the total annotation time}. We release our \textit{soft} labels in a new dataset, \texttt{CIFAR-10S}. 

\section{Related Work}
% We connect our work training with soft labels and eliciting uncertainty estimates from annotators. 
Training on soft instead of hard labels can improve robustness and generalization~\cite{pereyra2017regularizingEnt, muller2019labelSmoothHelp}. Soft labels have been constructed using smoothing mechanisms~\cite{labelSmoothing}, auxiliary teacher networks as in knowledge distillation~\cite{knowledgeDistill, gou2021knowledgeSurvey}, and aggregate human annotations~\cite{Sharmanska_2016_CVPR, peterson2019human, imagentRecht, Uma_Fornaciari_Hovy_Paun_Plank_Poesio_2020, gordon2021disagreement, gordon2022jury, uma2022scaling, koller2022beyondOneHot}. While the first two methods have led to significant advances in model performance, hand-crafted or learned soft labels often rely on hard labels, which tend to be impoverished representations of human precepts over datapoints. We therefore focus on the third approach, learning with soft labels derived from human annotations.
%though we note that the methods can be combined~\cite{Uma_Fornaciari_Hovy_Paun_Plank_Poesio_2020,uma2022scaling}.

% unsure whether we should decompose the above, or show how they're connected -- see Yuan_2020_CVPR_KD_LS

% Soft labels can be made using annotator disagreements in elicited labels~\cite{nguyen2014learning}. 
% \citet{nguyen2014learning} use discretized soft labels in clinical data.

% \citet{nguyen2014learning} form soft
% labels from experts’ subjective scales in a binary classification in a clinical data setting (equivalent to annotator confidence). 

For image data, \citet{peterson2019human} construct soft labels -- but do so by aggregating annotators' hard labels for~\texttt{CIFAR-10} -- and significantly improve classifier robustness. \citet{Uma_Fornaciari_Hovy_Paun_Plank_Poesio_2020} and \citet{uma2022scaling} extend aggregation-based soft labels to domains beyond image classification and study the incorporation of other forms of ``softness,'' such as temperature scaling, to enhance performance. Other works also primarily focus on aggregating \textit{hard} labels from individual annotators, which have since been used for several applications, from benchmarking the reliability of foundation models \citep{plex}, to informing human-machine teaming \citep{babbar2022utility, straitouri2022provably}, and expanding the empirical understanding of the impact of labels on performance \citep{wei2022aggregate,schmarje2022one}. While these labels are very valuable for the community, they are often touted as representing human ``label uncertainty'' \citep{plex}. Although such labels do capture some form of uncertainty, the picture is incomplete in only covering ambiguity \textit{across} humans, and does not capture each individual's uncertainty. Several excellent works have looked at collecting annotator confidence to address this limitation; however, in such cases, uncertainty is expressed over only the most probable label or in binary classification settings \citep{branson20questions, nguyen2014learning,song2018active, imagenetReaLH, steyvers2022bayesian}. In contrast, we believe we are the first to train using rich soft labels elicited \textit{directly} from annotators by requesting probabilistic judgments per annotator over multiple classes. 

%multi-class problems.%, and one of the first to employ such an approach more broadly.

% \citet{nguyen2014learning} form soft labels from experts' subjective uncertainty in a binary classification clinical data setting. 
% \citet{peterson2019human} also construct soft labels, now in a multi-class setting -- but instead -- do so aggregating annotators' hard labels for~\texttt{CIFAR-10} and  significantly improve the robustness of image classifiers. \citet{Uma_Fornaciari_Hovy_Paun_Plank_Poesio_2020} and \citet{uma2022scaling} extend this aggregation-based soft labels to domains beyond image classification and study the incorporation of other forms of ``softness,'' such as temperature scaling, to enhance performance. Existing works primarily focus on aggregating \textit{hard} labels from individual annotators. In contrast, we believe we are the first to train using rich soft labels elicited \textit{directly} from annotators by requesting probabilistic judgments per annotator for multi-class problems.%, and one of the first to employ such an approach more broadly.

There is a plethora of existing works on eliciting uncertainty judgements about outcomes~\cite{uncertainJudgments,nguyen2014learning,Firman_2018_CVPR,bhatt2021uncertainty,steyvers2022bayesian,vodrahalli2022uncalibrated}; however, none explicitly considers using uncertainty estimates to craft a soft label over multiple categories for training.
The crowdsourcing literature has looked into the efficiency of collecting additional information from annotators~\cite{efficientElic, audioUncertainty}.
We explicitly ask annotators to provide probabilistic judgments about labels for images, and focus our comparison not just on obtaining a soft label with fewer annotators than~\cite{peterson2019human} but also directly using our labels to confer improved machine performance.

\section{Problem Setting} 

% [NOTE: Need to update notation to match the label space focus from Wed meeting.]

% We begin by laying the groundwork of our single-annotator soft labeling method against conventional annotation methods, and highlight how they change the label space which a model is trained to fit.

We focus on the $K$-way classification setting. We assume we have a dataset of $N$ images $\{x_1, x_2, ...., x_N\}$ and an associated set of $N$ labels $\{y_1, y_2, ...., y_N\}$, where $y_i$ is a vector in $[0,1]^K$ representing a distribution over $K$ labels.
When the label is the traditional one-hot vector  $y_i \in \{0,1\}^K \subseteq [0,1]^K$, we call this a hard label. 
Each image's ordinal label may have been decided on by a single annotator~\cite{passonneau2014benefits}, or a majority vote of multiple annotators~\cite{sheng2017majority}. In our framework, this label distribution $P(y_n | x_n)$ has all mass placed on a single class: 
$$P_\text{hard}(y_n = k | x_n) = \mathds{1}[y_n = k],$$
\noindent where $\mathds{1}[y_n = k]$ is an indicator variable of whether class $k$ has been assigned or not by a single annotator. 
However, this approach does not allow for the representation of annotator disagreements. As a result, others consider eliciting a single ordinal label from each of $M$ annotators, $y_n^m \in \{0,1\}^K$~\cite{peterson2019human,Uma_Fornaciari_Hovy_Paun_Plank_Poesio_2020}. This results in a distribution over labels:
$$P_\text{multi}(y_n = k | x_n) = \frac{1}{M} \sum_{m=1}^M \mathds{1}[y_n^m = k].$$

The result is a soft label where $y_n \in [0,1]^K$. These labels can also be smoothed via softmax.
Yet, in existing frameworks, each annotator does not have the power to express their distribution over labels. We therefore consider the case where we elicit $P(y_n | x_n)$ directly from a \textit{single} annotator. Here, each annotator specifies $p_k^m$, their own personal probability distribution $P_m$ over the labels. This allows us to aggregate all $M$ annotators' probability distributions to form an aggregate label distribution as follows:
$$P_\text{ours}(y_n = k | x_n) = \frac{1}{M} \sum_{m=1}^{M} p_k^m,$$
\noindent where $p_k^m \in [0,1]$ that the label $y_n = k$, assigned by annotator $m$. We enforce the result is a valid probability distribution with $\sum_{k=1}^{K}p_k^i = 1$. This framing recovers a single hard label if $M=1$ and the annotator places all their mass on one label. We recover the soft label from~\cite{peterson2019human} if all $M$ annotators place all mass on one label.

% Note, converting a distribution $P(y_n | x_n)$ to a label during training is straightforward, if we permit soft labels at training time~\cite{Uma_Fornaciari_Hovy_Paun_Plank_Poesio_2020}. 

%\textit{all but one $p_k^m = 0$). }

% \paragraph{Aggregating over Annotators' Uncertainty}

% We can \textit{aggregate} all annotators' probability distributions if we wish to have a single label distribution for training. 
% We consider an aggregation function $\psi$ that takes a set of annotators' probability and outputs a single distribution:
% $$P(y_n = k | x_n) = \psi(\{p_k^m\}_{m=1}^{M})$$

% For example, we can let $\psi$ be the mean and take an average of all annotators' probability distributions:
% $$P_\text{ours}(y_n = k | x_n) = \frac{1}{M^\prime} \sum_{m=1}^{M^\prime} p_k^m$$

% We can design this aggregation function $\psi$ however we choose. 

% \paragraph{Handling Under-Specified Distributions}

Lastly, we consider the case where we do not have complete access to all $K$ (or $K-1$, by virtue of the sum-to-one constraint of valid probability distributions) $p_k^m$ per annotator. In practice, an annotator may only specify their confidence over the top two most likely labels, or perhaps indicate some subset of the $K$ labels which are likely to have zero probability given the image. Therefore, the annotator only provides $K'$ probability estimates where $K' < K - 1$. In this case, we need some method which distributes the leftover probability mass over the remaining $K - K' - 1$ labels. We refer to ``completing'' these under-specified distributions as the problem of \textit{re-distribution}. 

To handle \textit{re-distribution}, we define a function $r$ which takes as input any elicited probabilities from the annotator, and outputs a length $K$ vector $\hat{p}_{m}$, representing the ``completed'' set of $K$ probabilities over the label space (where $\sum_{k=1}^{K}\hat{p}_k^m = 1$). We then let:

$$P_m(y_n = k | x_n) = r(\{p_j^m\}_{j=1}^{K'})_k = \hat{p}_{m}^k.$$

We consider various designs for $r$ in Sec. 4.2. The resulting distributions can then be aggregated to yield a single distribution per image.\footnote{While in this work, we only consider naive aggregation, we discuss in Section 6 how more sophisticated aggregation methods could be explored in the future.} We highlight in Fig. \ref{fig:overview} the differences between eliciting label distributions which place all mass on a single class (\texttt{CIFAR-10H}) versus the soft labels we elicit and aggregate here. 
%, and leave further exploration of $\psi$ for future work.  \todo{we consider multiple $r$ but just one $\psi$ here. so perhaps we should remove $\psi$ ref?}

\section{Eliciting Soft Labels from Annotators}
We now discuss how we collect our dataset, \texttt{CIFAR-10S}. To elicit soft labels from each annotator, we request: 

\begin{enumerate}
    \item The most probable label, with an associated probability
    \item Optionally the second most probable label, with an associated probability
    \item Any labels which the image is \textit{definitely not}
\end{enumerate}

The most probable and second most probable labels are selected via a radio button, whereas the selection of ``definitely not'' possible labels is marked through a checkbox to allow annotators to select multiple labels. Probabilities are entered in a text box and asked to be between $0$ and $100$. We do not require that probabilities sum to $100$ across the task, as we normalize after by using one of the elicitation practices of \citet{uncertainJudgments}. We explore spreading any remaining mass over the labels not marked as impossible. 

As we observe that an overwhelming proportion of \texttt{CIFAR-10H} images have mass on only two labels ($\approx$77.2\%), we ask participants to specify only the top two most probable labels and any that are definitely not possible.

\begin{figure}[htb]
  \begin{center}
  \includegraphics[width=1.0\linewidth]{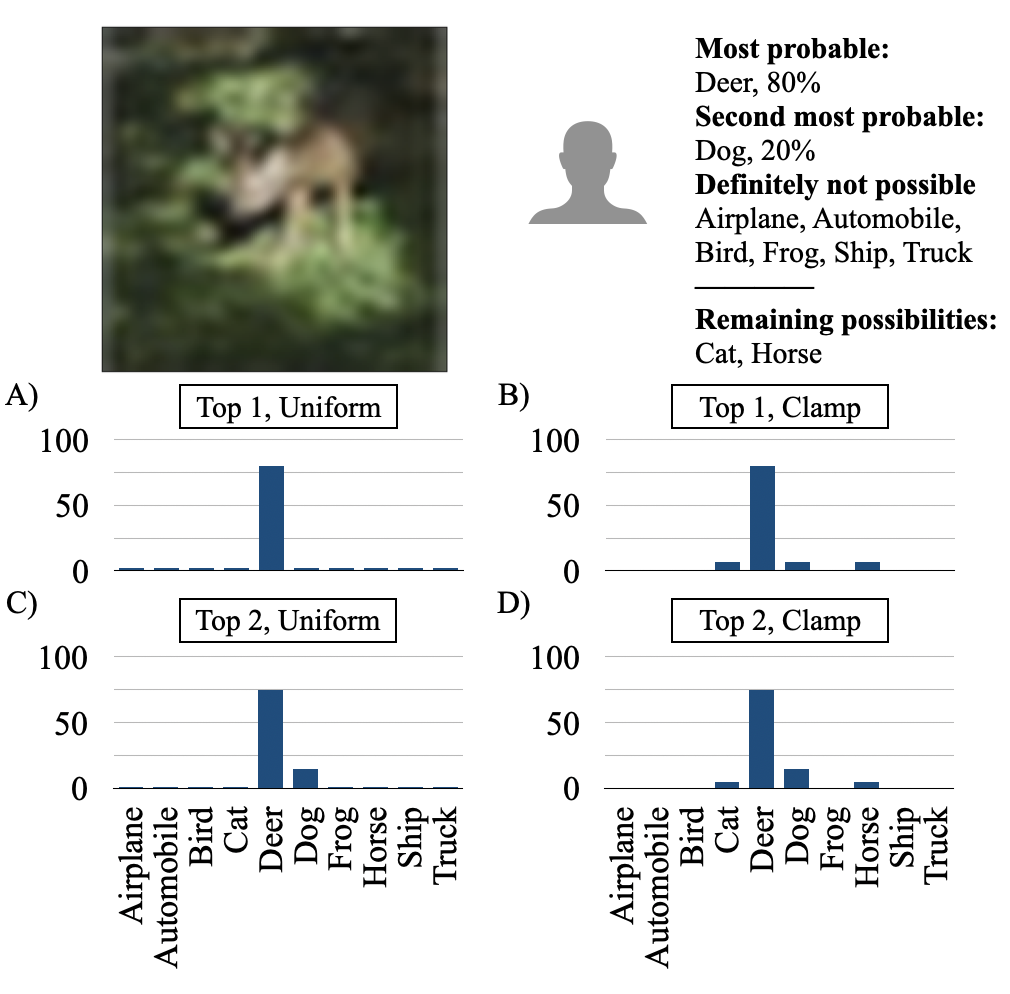}
      \caption{Depiction of constructed label varieties from the information elicited from a single annotator. Label type is depicted above the associated distribution. Note, possible labels are inferred by exclusions.}
        \label{fig:labelCreation}
  \end{center}
  \end{figure}
  %A) Top 1, Uniform, B) Top 1, Clamp, C) Top 2, Uniform, D) Top 2, Clamp.å

We additionally request annotators consider how \textit{other} annotators, specifically ``100 crowdsourced workers,'' may respond. Encouraging annotators to consider a \textit{third-person} perspective is reminiscent of Bayesian Truth Serum \citep{prelec2004bayesian} has been shown to encourage more representative responses \cite{efficientElic, oakley2010shelf}. Our interface is depicted in the Appendix.

\subsection{Setup} We recruited $N = 248$ participants on Prolific \cite{palan2018prolific}. Participants were recruited from the United States and required to speak English as a first-language. We identified 886 images with the highest entropy (entropy $\geq$ 0.25) from \texttt{CIFAR-10H} \cite{battleday2020capturing, peterson2019human} to best validate the efficacy of our approach on the ``hardest,'' and arguably most interesting cases. \citet{battleday2020capturing} found that only around 30\% of the images had high inter-annotator disagreements. However, we included three images with low entropy (entropy $\leq$ 0.1) under \texttt{CIFAR-10H} within each batch\footnote{With the exception of two of the batches of the $40$ batches that contain all higher entropy images due to randomization.} to ensure a sufficient diversity of ambiguity was shown to each participant. 

We  follow \citet{battleday2020capturing} in \textit{up-sampling} each image to a resolution of 160x160 using Lanczos-upsampling. While this reduces the ambiguity in the traditionally low-resolution \texttt{CIFAR-10} images, we aim to benchmark our method against \cite{peterson2019human} as closely as possible and hence follow their transformation. 
%-- rather than say, asking for the top three most probable categories, which may be less information-rich in this setting. 

Each participant sees a batch of $27$ images, where two images are repeated as checks for attention and consistency. The order of labels and images was shuffled across participants. To align with local regulations, annotators are paid a base rate of $\$8$/hr with a possible bonus up to a rate of \$9/hr. 

% \paragraph{Exclusion Criteria} 
We exclude any participant who did any of the following more than twice: (i) specified a probability outside of the range requested, 0 to 100; (ii) expressed that their own most probable or second most probable labels were also definitely not possible; 
or (iii)  failed to specify any probability for their most probable label. 
For participants who made such errors only once, we only rule out those who provided low-quality responses by excluding those who had an accuracy against the \texttt{CIFAR-10} hard labels less than 75\%, the threshold used in \cite{battleday2020capturing}. Our 75\% accuracy exclusion threshold is only applied for annotators who made one of the above errors. We never exclude by accuracy alone in an effort to maintain diversity of percepts collected, as accuracy assumes \texttt{CIFAR-10} labels are ground truth. 

% \paragraph{Metrics} After applying our exclusion criteria, we yield a total of $125$ participants, who obtain an average accuracy against \textt

\subsection{Constructing Soft Labels} Our elicitation yields multiple pieces of information (first and second most probable labels with specified probabilities, and labels which are deemed to have zero probability) which we can use -- or ignore -- when forming a soft label. We explore several varieties of soft label constructions. 

\paragraph{How to Redistribute Extra Mass?} A central question in our elicitation scheme is how to distribute any mass which is left unspecified; for instance, if an annotator marks ``truck'' as the most probable class with probability 70\% and ``automobile'' as the second most probable class at 20\% likely, there is 10\% of mass remaining that conceivably could be spread onto other classes. We consider two forms of redistribution in this work: 1) \textbf{uniform} redistribution whereby the remaining mass is spread equally over the remaining classes, or 2) \textbf{clamp} which uses the ``definitely not'' elicitation to spread the remaining mass equally over only those classes which the annotator did not specify as zero probability.
%Future work can consider redistribution that considers the semantic distance between labels~\cite{word2vec}.%In the clamp setting, we assume the remaining, unspecified classes are mentally \textit{plausible}.

If an annotator specifies 100\% of the mass over the top one or two labels but only selects a subset of the remaining labels as definitely not possible, then we posit that the annotator views the unselected classes not having zero probability. Thus, we maintain a small portion of mass $\gamma$ to be spread over the remaining classes. $\gamma$ is selected via a held-out set, as discussed in Section 5.1.  We do not apply this procedure in the \textit{uniform} redistribution setting,  as there we assume no access to the ``definitely not'' information.

\paragraph{Label Varieties} We have $2$ x $2$ possible soft label construction methods: \{most probable only, most probable and second most probable\} x \{redistribute uniformly, redistribute via clamp\}. We use the notation T1 to specify if only the most probable class and its associated probability is used, and T2 if we include information about both the most probable and second most probable categories. We also refer to the redistribution approaches as ``clamp'' or ``unif'' following the definitions above. The label that uses \textit{all} elicited information is T2 Clamp, which is the label set we refer to as \texttt{CIFAR-10S}. All soft labels, regardless of variety, are normalized to sum to one. Examples of  constructed labels from a single annotators' response are shown in Fig. \ref{fig:labelCreation}.

\subsection{Comparing \texttt{CIFAR-10H} and \texttt{CIFAR-10S} Label Properties} We compare the structure of our elicited labels in \texttt{CIFAR-10S} to \texttt{CIFAR-10H} labels \cite{peterson2019human, battleday2020capturing}. The elicitation of \texttt{CIFAR-10H} is lossy because annotators may be less than $100\%$ sure about the hard label they are asked to provide. Consider if every annotator is $51\%$ sure an image is class $k$ and $49\%$ sure it is class $\ell$, then they will only provide class $k$ in the elicitation of \citet{peterson2019human}. For our setting, annotators can express their label probabilities directly. 

\begin{figure}[htb]
  \begin{center}
  \includegraphics[width=0.9\linewidth]{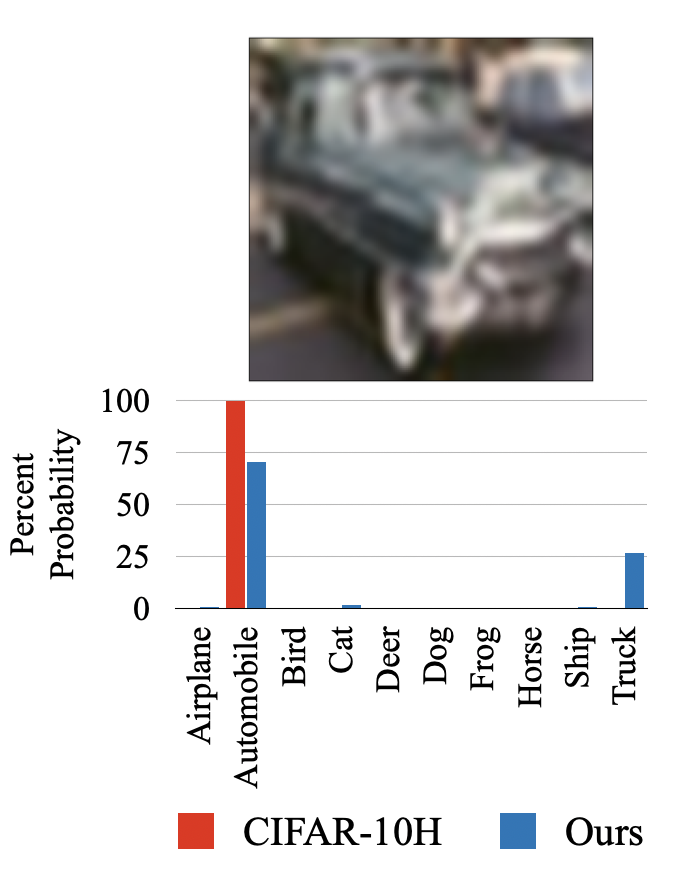}
      \caption{Our soft variant of a \texttt{CIFAR-10H} hard label captures inter-class similarities (i.e., trucks and automobiles).}
  \label{fig:hardVSoft}
  \end{center}
  \end{figure}

% \texttt{CIFAR-10H} is asking annotators the wrong question: asking annotators to provide a hard label, when in practice, annotators may be using an ``internal soft label'' \cite{} fails to capture annotator uncertainty.  

% \paragraph{Illustrative Example} We start with an illustrative example of how our labels have the potential to fundamentally differ. Consider an image where annotators 51\% sure the image is class $k$ and 49\% sure the image is class $\ell$. Each annotators' hard labels will  be class $k$, so all mass in the aggregated label will be on $k$, resulting in a hard label. However, if annotators are empowered to express their uncertainty in the classes label -- for instance, by specifying what they deem the most likely and second most likely classes, along with their probabilities -- then the aggregated label will recover a label that more faithfully captures human uncertainty.

\begin{figure*}[htb]
  \begin{center}
  \includegraphics[width=1.0\linewidth]{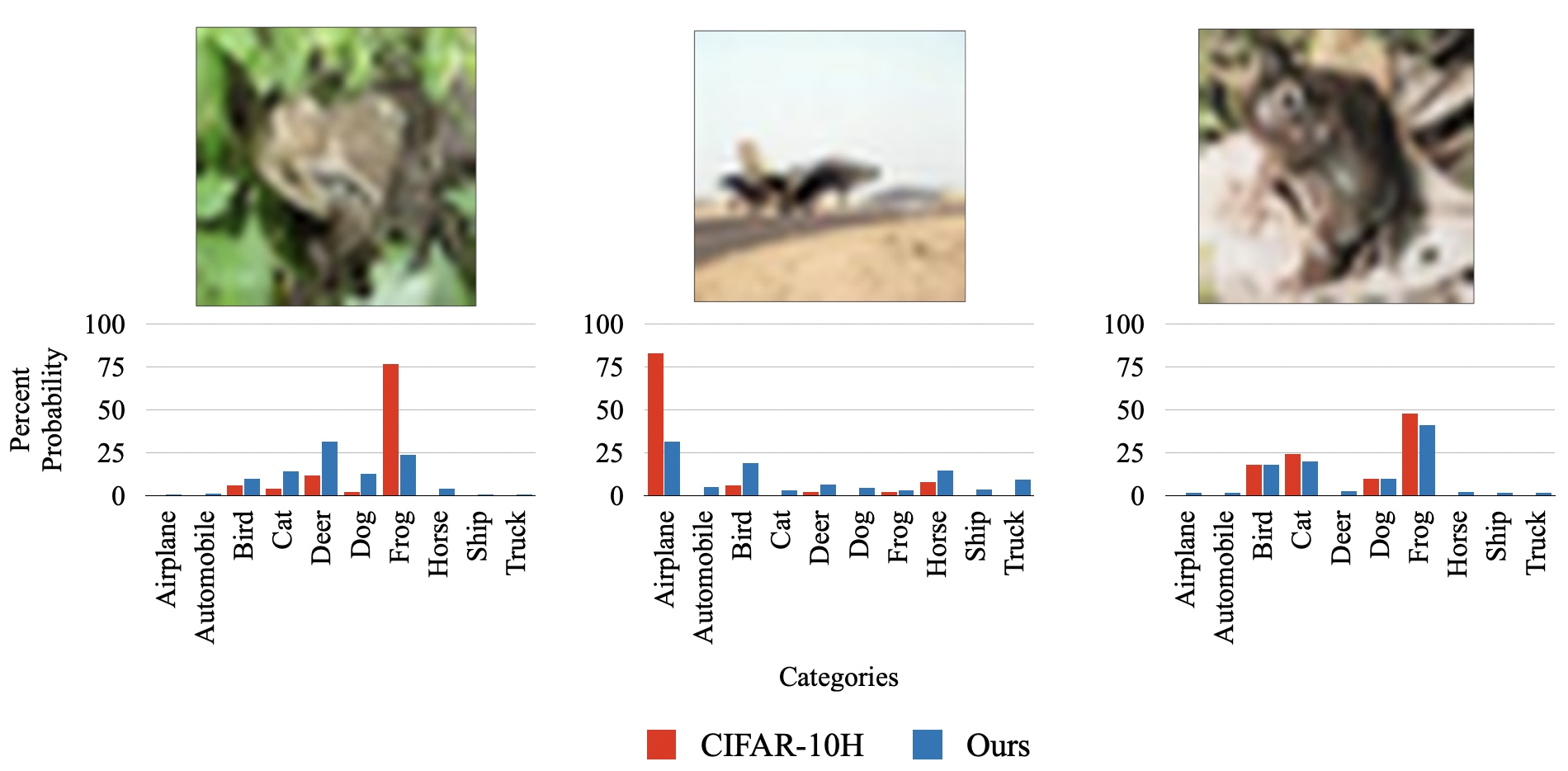}
      \caption{Comparison of our elicited labels against \texttt{CIFAR-10H}. From left to right: the images depict two examples with high Wasserstein distance between labels; then one example where we recover similarly rich, high entropy labels %ambiguous \texttt{CIFAR-10H} examples 
      from \textit{8.5}x fewer annotators. The \texttt{CIFAR-10} labels for these images are frog, airplane, and frog, respectively. Images depict what annotators actually saw; i.e., upsampling is applied following \citet{battleday2020capturing}.}
  \label{fig:compareLabels}
  \end{center}
  \end{figure*}

% \paragraph{Empirical Differences} 
% We see the emergence of similar labels in our data. 
% There are similar labels in \texttt{CIFAR-10H} and \texttt{CIFAR-10S}.
While \texttt{CIFAR-10H} labels may have nearly all mass on a single class, our elicitation yields labels which have mass spread across more classes. This not only captures some of the inherent ambiguity in an image, but has the potential to provide information about the inter-class similarity structure. For example, our annotators place mass jointly over ``automobiles'' \textit{and} the similar ``truck'' category, whereas a \texttt{CIFAR-10H} label may have all mass on the ``automobile'' category; see Fig. \ref{fig:hardVSoft}. We highlight additional examples of label differences in the Appendix. 
%Although we do not study inter-class similarity structure in this work, this direction is ripe for further inquiry.

% Examples are depicted in Fig. \todo{ref}. For instance, we notice that an image which may be majority-labeled as an ``automobile'' engenders our annotators to maintain some probability mass on other similar label categories like ``truck'' which could be possible. We reason information about which classes which co-vary as being marked probable can enable a learner to better capture inter-class similarities. 

% \begin{table}[htb]
% \centering
% \begin{tabular}{ccc}
% \hline
% M & Distance & Entropy Corr \\ \hline
% 1 & 0.033$\pm$0.002 & 0.348 \\
% 2 & 0.032$\pm$0.002 & 0.432 \\
% 3 & 0.030$\pm$0.002 & 0.498 \\
% 4 & 0.029$\pm$0.002 & 0.545 \\
% 5 & 0.029$\pm$0.001 & 0.567 \\
% 6 & 0.028$\pm$0.001 & 0.596 \\ \hline
% \end{tabular}
% \caption{Comparison of Wasserstein distance and entropy correlation between our labels and CIFAR-10H.}
% \label{tab:compDistEnt}
% \end{table}

  \begin{figure}[htb]
  \begin{center}
  \includegraphics[width=1.0\linewidth]{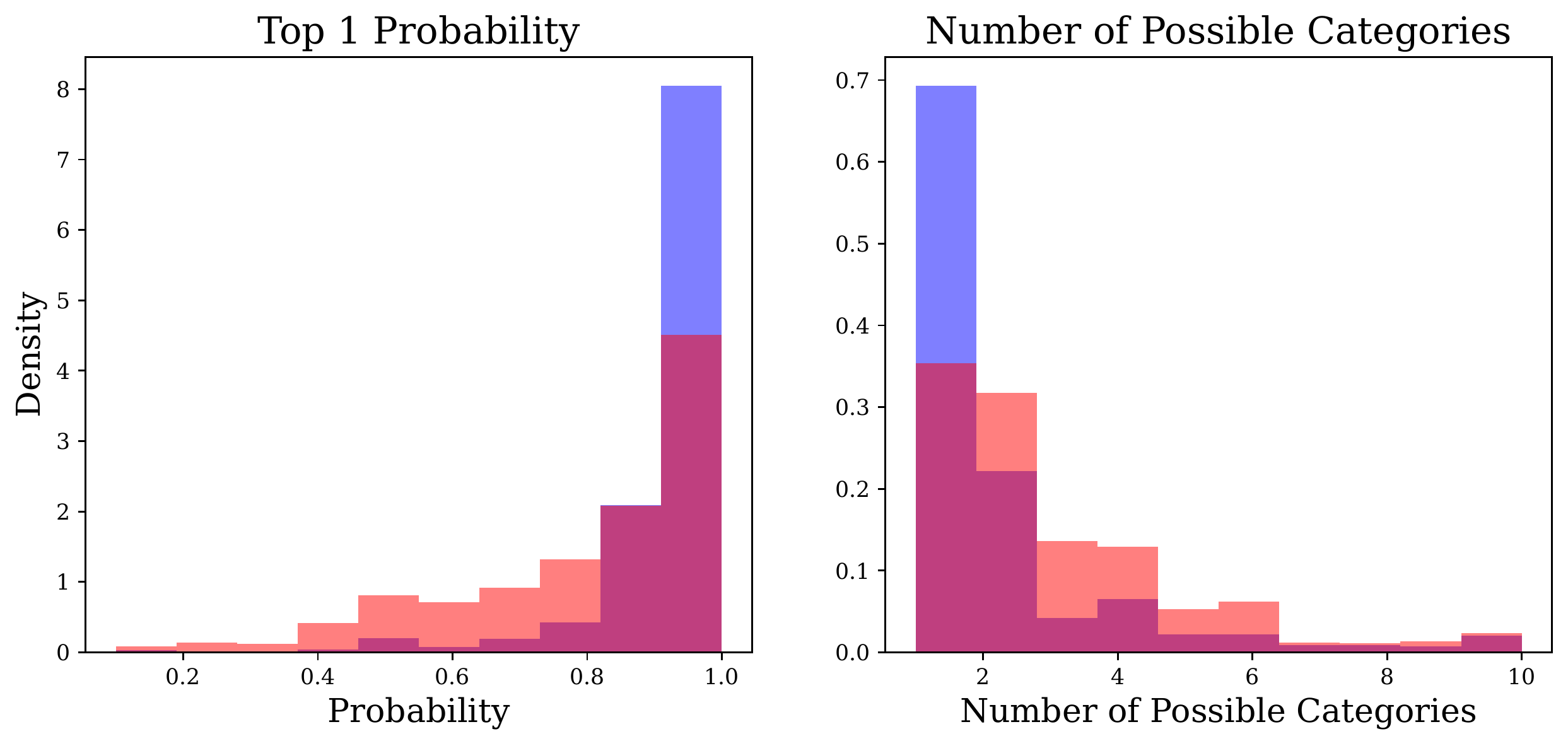}
      \caption{Amount of probability placed on the most probable label and the number of ``possible'' labels (i.e., all those which are not selected as ``Definitely Not'' possible) in the T2 Clamp soft labels formed from \texttt{CIFAR-10S} annotators tracks \texttt{CIFAR-10H} label entropy (high = red, low = blue).} 
    \label{fig:entropyCompare}
  \end{center}
  \end{figure}

% One may ask, however, whether our framework is \textit{over-}encouraging softness? As we sprinkled in low entropy images for most batches, we can compare the forms of the soft labels elicited for low versus high entropy images. We see in Fig. \todo{ref} that annotators place higher confidence on their most probable label and indicate fewer alternative possible labels (determined as all labels not selected as ``impossible'') for images which are low entropy (according to \texttt{CIFAR-10H}) than high. These data suggest that our elicitation framework is sensible, i.e., the set-up does not appear to be over-encouraging non-sensical annotator uncertainty, but instead, capturing semantically-meaningful ambiguity across categories. %Though we acknowledge that more work is needed to fully elucidate the form of soft labels gleaned from our elicitation framework when examples have little ambiguity. 

While our elicitation scheme yields fundamentally different labels for some images, we find our method produces remarkably similar labels to those of \texttt{CIFAR-10H}. Specifically, when considering T2 Clamp, the entropy of our labels has a Pearson's $r$ correlation coefficient of 0.596, and an average Wasserstein distance of 0.028 to \texttt{CIFAR-10H}. This is encouraging, as we are able to recover much of the richness of \texttt{CIFAR-10H} labels \textit{and more} from approximately 8.5x fewer annotators (an average of $M = 51$ per image vs. approximately $6$ in our dataset). We depict this in Fig. \ref{fig:compareLabels}. The amount of probability mass placed on the most probable category and the number of ``impossible'' labels track the \texttt{CIFAR-10H} label entropy (see Fig \ref{fig:entropyCompare}), which suggests that our labeling interface does yield sensible soft labels for clearer (low entropy) images.

\begin{table*}[htbp]
    \centering
    \begin{tabular}{ccccc} 
    \toprule
     & Label Type       & CE             & Calibration    & FGSM Loss       \\
    \midrule
    \multirow{5}{*}{\STAB{\rotatebox[origin=c]{90}{ \texttt{CIFAR-10H}}}} & Hard Labels      & 2.026$\pm$0.18 & 0.277$\pm$0.01 & 15.455$\pm$7.5  \\
 & Random Labels    & 1.770$\pm$0.16  & 0.226$\pm$0.02 & 13.476$\pm$1.39 \\
&  Uniform Labels   & 1.599$\pm$0.13 & 0.203$\pm$0.02 & 10.199$\pm$3.82 \\
 & CIFAR-10H        & \textbf{1.325$\pm$0.07} & \textbf{0.201$\pm$0.01} & \textbf{8.750$\pm$1.80}    \\
 & Ours (T2, Clamp) & 1.369$\pm$0.07 & 0.203$\pm$0.01 & 8.872$\pm$1.63  \\
     \midrule
     \multirow{5}{*}{\STAB{\rotatebox[origin=c]{90}{ \texttt{CIFAR-10S}}}} & Hard Labels      & 4.460$\pm$0.49  & 0.425$\pm$0.09 & 15.782$\pm$4.67 \\
     & Random Labels    & 3.093$\pm$0.53 & 0.353$\pm$0.05 & 10.697$\pm$4.14 \\
     & Uniform Labels   & 2.923$\pm$0.26 & \textbf{0.311$\pm$0.06} & 11.768$\pm$5.18 \\
     & CIFAR-10H        & \textbf{2.558$\pm$0.16} & 0.313$\pm$0.03 & \textbf{8.416$\pm$1.64}  \\
     & Ours (T2, Clamp) & 2.591$\pm$0.19 & 0.324$\pm$0.02 & 9.116$\pm$1.63  \\
    \bottomrule
    \end{tabular}

    \caption{Comparing performance when training on labels aggregated from humans' annotations ($M = 51$ of \texttt{CIFAR-10H} annotators, $M=6$ of ours). Our soft labels here utilize all elicited information (i.e., T2 Clamp). Different labels are considered over $900$ examples. Performance is evaluated over $3000$ heldout \texttt{CIFAR-10H} labels (top) and $100$ heldout labels from our collection (the T2 Clamp variant of \texttt{CIFAR-10S}, bottom). 95\% confidence intervals are included. 5 seeds are run each for three models (ResNet-34A, ResNet-50, VGG-11) and averaged. Bold indicates best performance (lower is better for all metrics).}
    \label{tab:mAll_compare}
\end{table*}

% \begin{table*}[htb]
% \centering
% \begin{tabular}{llllllll}
% \toprule
%  Label Type               & Time   & Acc           & CE            & FGSM Loss      & PGD Loss       & Calibration Error   & Entropy Corr

%  \\
% \midrule
%  Hard Labels              &    -    & 0.64$\pm$0.01 & 2.25$\pm$0.04 & 16.72$\pm$0.23 & 27.62$\pm$0.33 & 0.3$\pm$0.01        & 0.14$\pm$0.02         \\
%  CIFAR-10H               &   306s     & \textbf{0.66$\pm$0.01} & \textbf{1.38$\pm$0.02} & \textbf{9.45$\pm$0.12}  & \textbf{17.17$\pm$0.19} & \textbf{0.2$\pm$0.01}        & \textbf{0.21$\pm$0.01}         \\
%  Ours (T2, Clamp) &    \textbf{180s}    & 0.65$\pm$0.01 & 1.41$\pm$0.02 & 9.52$\pm$0.14  & 19.57$\pm$0.21 & \textbf{0.2$\pm$0.01}        & \textbf{0.21$\pm$0.01}         \\
% \bottomrule
% \end{tabular}
%     \caption{Training on 1) original \texttt{CIFAR-10} hard labels, 2) aggregated per-annotator hard labels from \texttt{CIFAR-10H} (avg M = 51 annotators per image), versus 3) the $M=6$ aggregated per-annotator soft labels we collect in \texttt{CIFAR-10S}. Our soft labels here utilize all elicited information from annotators (i.e., T2 Clamp).}
%     \label{tab:mAll_compare}
% \end{table*}

\paragraph{\textit{Takeaways}} Our elicitation approach enables annotators to express their distribution over image labels, approximating and expanding on the richness of \texttt{CIFAR-10H} from far fewer annotators. Even in cases where annotators may agree on the most likely image label, our approach -- which enables annotators to express their probability judgments over \textit{possible} other categories -- yields labels which potentially better represent the distribution over labels.

\begin{table*}[htbp]
    \centering
    \begin{tabular}{ccccc} 
    \toprule
     & Label Type       & CE             & Calibration    & FGSM Loss       \\
    \midrule
    \multirow{2}{*}{\STAB{\rotatebox[origin=c]{90}{ \texttt{10H}}}} & CIFAR-10H       & 1.293$\pm$0.08 & 0.194$\pm$0.01 & 8.577$\pm$1.91 \\
  & Ours (T2, Clamp) & \textbf{1.281$\pm$0.06} & \textbf{0.184$\pm$0.01} & \textbf{8.406$\pm$1.75} \\
     \midrule
     \multirow{2}{*}{\STAB{\rotatebox[origin=c]{90}{ \texttt{10S}}}} & CIFAR-10H       & 2.459$\pm$0.21 & 0.311$\pm$0.02 & \textbf{8.334$\pm$1.75} \\
  & Ours (T2, Clamp) & \textbf{2.355$\pm$0.14} & \textbf{0.297$\pm$0.03} & 8.405$\pm$1.59 \\
    \bottomrule
    \end{tabular}

    \caption{Training with de-aggregated labels; on each batch, a single humans' label is used as supervisory signal from a pool of $M$ humans ($M=51$ \texttt{CIFAR-10H}; $M=6$ ours).}
    \label{tab:deaggFull}
\end{table*}

\section{Evaluating the Efficacy and Efficiency of Learning with Per-Annotator Soft Labels} 

We investigate how learning over our elicited soft labels compares against learning over labels drawn from \texttt{CIFAR-10H}. \texttt{CIFAR-10H} labels have previously been shown to confer generalization benefits over several other soft labeling approaches, such as class-level confusion-based smoothing and knowledge distillation \cite{peterson2019human}.
%-- hence why we focus specifically on \texttt{CIFAR-10H}. 
We focus on performance when constructing labels from \textit{sub-samples} of annotators. This enables us to investigate the \textit{annotator efficiency} of our soft label elicitation. Additionally, we study the impact of constructing labels from subsets of elicited data.
%-- as discussed in Section \todo{4.2}. 
We also consider performance differences in light of the \textit{total time} elicitation takes. 

% Learning over \texttt{CIFAR-10H}

% We next investigate the benefits of our elicited soft labels on model generalization, robustness, and calibration. We focus our comparisons against the CIFAR-10H labels, which have previously been shown to confer numerous advantages to models trained on them compared to both conventional one-hot CIFAR-10 labels, as well as other labeling methods such as class-level soft labels, a knowledge distillation, and mixup \cite{peterson2019human, Uma_Fornaciari_Hovy_Paun_Plank_Poesio_2020}. We investigate the comparative performance of learning with aggregated versions of our soft labels -- as well as several label variants derived from our full elicitation, e.g., just using the most probable category with its elicited probability -- against labels constructed from a sub-sampling of the annotators' hard labels that make up CIFAR-10H. 

\subsection{Setup}

% \paragraph{Data} We follow \citet{Uma_Fornaciari_Hovy_Paun_Plank_Poesio_2020} by training over a 70/30 split of \texttt{CIFAR-10H}. As there are $10,000$ images in \texttt{CIFAR-10H}, we use $3,000$ images as our evaluation set. Validation is performed over a 10\% \texttt{CIFAR-10} training set. As we do not have labelings for  all the images in \texttt{CIFAR-10H}, we train on hard labels for any image for which we have not elicited labelings: this ensures a fair comparison between our soft labels and those of \texttt{CIFAR-10H}. \textbf{Since our models are trained on labels from our data but tested on labels from~\cite{peterson2019human}, we expect to perform worse than had we tested on our own labels.}\todo{revise section now that we eval on \texttt{10S}}
% Our comparison against \texttt{CIFAR-10H} relates to the labelings of the selected $1,000$ images. 

\paragraph{Setup} Our model and training procedures follow \citet{Uma_Fornaciari_Hovy_Paun_Plank_Poesio_2020}, %. We employ procedures from \citet{Uma_Fornaciari_Hovy_Paun_Plank_Poesio_2020}, 
as they explore learning with \texttt{CIFAR-10H} labels and explicate a clear, standardized learning procedure. We employ the same ResNet-34A~\citep{he2016deep} with the same weight decay (1e-4) and learning rate scheduling: we start with a learning rate of $0.1$ and drop by a factor of 1e-4 after epoch $50$ and again at $55$. To compare the utility of our soft labels during learning across a range of settings, we also consider two additional architectures not included in \citet{Uma_Fornaciari_Hovy_Paun_Plank_Poesio_2020}: VGG-11~\citep{simonyan2014very} and ResNet-50~\citep{resnet}. Starting learning rates are selected using a validation set of \texttt{CIFAR-10} ($0.1$ for VGG-11 and $0.4$ for ResNet-50 from $\{0.001, 0.01, 0.1, 0.2, 0.3, 0.4, 0.5\}$. All models are trained from scratch for a total of $65$ epochs and optimize a cross-entropy objective. Experiments are run over $5$ seeds, unless otherwise noted. A redistribution factor of $\gamma = 0.1$ is used to spread extra mass, selected via the same validation procedure from $\{0.0, 0.01, 0.05, 0.1, 0.2, 0.3, 0.4\}$. 

We follow the same 70/30 split used in \citep{Uma_Fornaciari_Hovy_Paun_Plank_Poesio_2020}. However, as we have fewer \texttt{CIFAR-10S} labels than \texttt{CIFAR-10H}, to ensure a fair comparison, we consider just training on the \texttt{CIFAR-10H} labels for which we have our soft label versions. Note, we always hold out $100$ of our labels to ensure we can evaluate against some variant of our soft labels. Thus, we are considering the variation in performance conferred by changing 900 labels. For the remaining examples in the 6,100 -- we use a hard version, i.e., the original \texttt{CIFAR-10} label.

% \paragraph{Model and Training} Our model and training procedures follow \citet{Uma_Fornaciari_Hovy_Paun_Plank_Poesio_2020}, %. We employ procedures from \citet{Uma_Fornaciari_Hovy_Paun_Plank_Poesio_2020}, 
% as they explore learning with \texttt{CIFAR-10H} labels and explicate a clear, standardized learning procedure. We similarly employ a ResNet-34A \cite{he2016deep} with a weight decay of 1e-4 and follow their learning rate scheduling: we start with a learning rate of $0.1$ and drop by a factor of 1e-4 after epoch $50$ and again at $55$. \todo{add other model note} We train for a total of $65$ epochs and employ a cross-entropy training objective. We run each experiment over $10$ seeds. A redistribution factor of $\gamma = 0.1$ is used to spread extra mass, which was selected using a validation set of \texttt{CIFAR-10} from $\gamma \in \{0.0, 0.01, 0.05, 0.1, 0.2, 0.3, 0.4\}$. 

\begin{table*}[htbp]
    \centering
    \begin{tabular}{cc|ccc|ccc} 
        \toprule
  \multicolumn{2}{c}{} & \multicolumn{3}{c}{M=2} & \multicolumn{3}{c}{M=1} \\

     & Labels       & CE             & Calib    & FGSM & CE             & Calib    & FGSM     \\
    \midrule
    \multirow{2}{*}{\STAB{\rotatebox[origin=c]{90}{ \texttt{10H}}}} & 10H & 1.71$\pm$0.15 & 0.25$\pm$0.02 & 14.16$\pm$0.67 & 2.17$\pm$0.11 & 0.29$\pm$0.01 & 19.20$\pm$0.86 \\
 & Ours & \textbf{1.49$\pm$0.08} & \textbf{0.22$\pm$0.01} & \textbf{11.57$\pm$0.44} & \textbf{1.58$\pm$0.08} & \textbf{0.23$\pm$0.01} & \textbf{12.57$\pm$1.29} \\
    \midrule
     \multirow{2}{*}{\STAB{\rotatebox[origin=c]{90}{ \texttt{10S}}}} & 10H & 3.37$\pm$0.42 & \textbf{0.34$\pm$0.08} & 13.13$\pm$0.7 & 4.49$\pm$0.31 & 0.45$\pm$0.05 & 17.85$\pm$0.68 \\
 & Ours & \textbf{2.88$\pm$0.17}  & 0.36$\pm$0.04 & \textbf{11.39$\pm$0.59} & \textbf{2.90$\pm$0.31} & \textbf{0.38$\pm$0.06} & \textbf{12.13$\pm$1.31} \\
    \bottomrule
    \end{tabular}

    \caption{Investigating model performance when fewer $M$ annotators are assumed to provide labels. Training labels are sampled per batch instead from a pool of $M=2$ or $M=1$ annotators.}
    \label{tab:few_m_compare}
\end{table*}

\subsubsection{Evaluation Data} Selection of data used to evaluate models is important to faithfully benchmark performance; however, typical evaluation datasets like the \texttt{CIFAR-10} test set are rife with annotation errors \citep{northcutt2021pervasive} and as discussed throughout this work, if hard labeled, do not adequately capture human uncertainty. We instead use heldout aggregate soft labels from \texttt{CIFAR-10H} and our \texttt{CIFAR-10S} as test sets. While humans are of course not always correct in their annotations themselves, nor calibrated in their confidence, these labels serve to better measure whether models handle example ambiguity. Note, as the labels we collect are enriched to be more ambiguous (see Section 4.1), at present, \texttt{CIFAR-10S} could be an inherently challenging evaluation set.

% When interpreting results, we note that while elicitation approach \texttt{CIFAR-10H} may converge to different label distributions than ours, we acknowledge that at present, \texttt{CIFAR-10H} is a more stable dataset given annotator quality is more tightly controlled \citep{battleday2020capturing}. We believe when scaled, our labels however could serve as a valuable benchmark; here, as we have few labels collected, we only maintain $100$ held-out (which are re-sampled per seed). We note that as the labels we collect are enriched to be more ambiguous, \texttt{CIFAR-10S} is a naturally challenging evaluation set.

\paragraph{Metrics} No single metric captures all the qualities we wish to obtain in a trustworthy model \cite{thomas2022relianceMetrics}. We therefore consider a suite of metrics, focused on generalization, calibration, and robustness. 

\begin{enumerate}
    \item Generalization: we measure generalization using Cross Entropy (CE) over the soft labels, which allows us to capture whether the models' full distribution over the $K$ categories is sensible: $\frac{1}{N}\sum_{1}^{N}\sum_{1}^{K}P_\text{eval}(y_n=k|x_n)\log(f_\theta(x_n)_k)$, where $P_\text{eval}$ is the label distribution derived from the human soft labels, drawn from either \texttt{CIFAR-10H} or our \texttt{CIFAR-10S} collection, and $f_\theta(x_n)_k$ is the probability assigned by our model $f$ (parameterized by $\theta$) to the $k$-th category on the $n$-th input. 
    \item Calibration: model calibration is scored using the RMSE adaptive-binning method used by \citet{hendrycks2022pixmix} to measure whether models' predictive distributions match their ``correctness''. Here, ``correct'' is based on: $\argmax_{k \in \{1,...,K\}}P_\text{eval}(y_n=k|x_n)$.  
    \item Robustness: loss after a Fast Gradient Sign Method (FGSM) attack is used to measure models' robustness to an adversarial attack \citep{goodfellowAdversarial}.\footnote{We also considered the multi-step Projected Gradient Descent, PGD~\cite{kurakin2016adversarial} attack; however, the metric was unstable and warrants further investigation.} Attack strength is run at an $\ell_\infty=4$ bound following \citet{peterson2019human}. 
\end{enumerate}

% \paragraph{Evaluation Measures} We care principally about model generalization, calibration, and robustness. However, as noted in \cite{thomas2022relianceMetrics}, no single metric captures the qualities we seek to obtain. Hence, we consider a battery of tests, including and beyond conventional accuracy against the most probable label. Specifically, we measure cross-entropy against the heldout \texttt{CIFAR-10H} soft labels, robustness to adversarial attacks, namely Fast Gradient Sign Method, FGSM~\cite{goodfellow2014explaining}\footnote{We explored a multi-step adversarial attack, specifically, Projected Gradient Descent, PGD~\cite{kurakin2016adversarial}, however, results were inconclusive \todo{edit?}.}, and RMSE calibration error with adaptive binning following \cite{hendrycks2022pixmix}. Both robustness attacks are employed using an $\ell_\infty=4$ bound with the PGD attack being run for $10$ iterations. These settings match those used in \citet{peterson2019human}. \todo{update}

\paragraph{Annotation Time} We include estimated total annotation time for each labeling scheme. We let total annotation time equal $M \times t_\text{per}$, where $M$ is the number of annotators being aggregated per image and $t_\text{per}$ is the estimated time taken per annotator to obtain that label type. Our elicitation takes a median of $32$ seconds for annotators to provide the most probable label with an associated probability, optionally the second most probable label with a probability, and any label which are definitely not perceived as the image category. This entails five different inputs from an annotator. As we do not have access to the time taken for each input, we assume that each takes roughly the same amount of time. We assign an estimated $6.4$ seconds to the elicitation time per input. We compute the median amount of time taken for \texttt{CIFAR-10H} annotators over the same images using their released raw annotation data, which comes to approximately $t_\text{per}=1.8$ seconds per image. Note, this value does not account for the training phase that \citet{battleday2020capturing} used per annotator, which could have increased total costs. However, even so, we acknowledge that comparatively high time costs of our elicitation scheme are a signification limitation of our work at present; designing more efficient soft label elicitation is a promising direction for future work.

\subsection{Learning with Soft Labels} 
We first compare our naively aggregated per-annotator soft labels using all information elicited from annotators (i.e., T2, Clamp) against the \textit{complete} aggregate labels from \texttt{CIFAR-10H}, i.e., labels formed from all of their approximately 51 labelers. As noted in Section 5.1, we use heldout \texttt{CIFAR-10H} and \texttt{CIFAR-10S} labels as a proxy for ``test truth'' when evaluating. We benchmark performance against training on conventional \texttt{CIFAR-10} hard labels, as well as training on random and uniform labels. A discussion of label smoothing is included in the Appendix.

As shown in Table \ref{tab:mAll_compare}, we find that, even from approximately 8.5x fewer annotators than \texttt{CIFAR-10H}, our per-annotator soft labels endow the learned classifier with performance comparable to that obtained by using the \texttt{CIFAR-10H} labels. %nearly comparable performance with respect to \texttt{CIFAR-10H} labels.

However, prior work has indicated that training on \textit{separated}, de-aggregated labels could yield better performance when there are few, noisy annotators \citep{wei2022aggregate}. \citet{peterson2019human} similarly found learning on de-aggregated labels was sometimes advantageous, which they hypothesize is from more varied gradient information. As such, we next compare model performance when trained on de-aggregated labels. That is, at each batch, a \textit{single} annotator's label (from the pool of $M$) is sampled for use as the supervisory signal. In this setting, we see in Table \ref{tab:deaggFull} that our \texttt{CIFAR-10S} labels confer substantial benefits over \texttt{CIFAR-10H} across nearly all metrics, despite having $8.5$x fewer annotators. As models trained on de-aggregated labels enjoyed better performance across both datasets, we use the de-aggregated setting for all further experiments. 

%Our labels took around 60\% of the total annotation time taken to acquire \texttt{CIFAR-10H} labels. 

% We evaluate how our labels compare against \texttt{CIFAR-10H} when accounting for 

% % \subsection{Comparing under Labels Derived from Few Annotators} 
% We now further probe how our elicitation is more \textit{annotator-efficient.} We construct labels using the \textit{same} total number of annotators across both labeling approaches. 
% %when aggregating annotators' hard labels or from our method of aggregating per-annotator soft labels.
% For each image, we sample two of the annotators in \texttt{CIFAR-10H} to aggregate, and compare the utility of learning over the aggregate labels against a similarly sub-sampled aggregation over two of our per-annotators' soft labels per image. When forming labels over few annotators, our method yields significant performance boosts, as seen in Table~\ref{tab:m2_compare}. We can go further and demonstrate in Table \ref{tab:m1_compare} that constructing training labels from \textit{just one} of our annotators achieves even larger robustness and downstream calibration benefits compared to learning from a single \texttt{CIFAR-10H} labeler. This is expected, as a single \texttt{CIFAR-10H} labeler is simply a one-hot, hard label. 

In Table \ref{tab:few_m_compare}, we subsample $M=2$ of the annotators in \texttt{CIFAR-10H} from which to construct a label per batch, and compare the utility of learning with said labels against a similarly sub-sampled version over two of our per-annotators' soft labels per image. We find that our labels provide a substantial boost along nearly all metrics -- and the gains of our method in the few annotator setting become even more apparent when considering access to only $M=1$ human. While this is expected, as a single \texttt{CIFAR-10H} labeler is simply a hard label, we demonstrate that \textbf{if one has access to only a single annotator, our label method provides the best training signal}.

\begin{figure*}[htb]
  \begin{center}
  \includegraphics[width=0.8\linewidth]{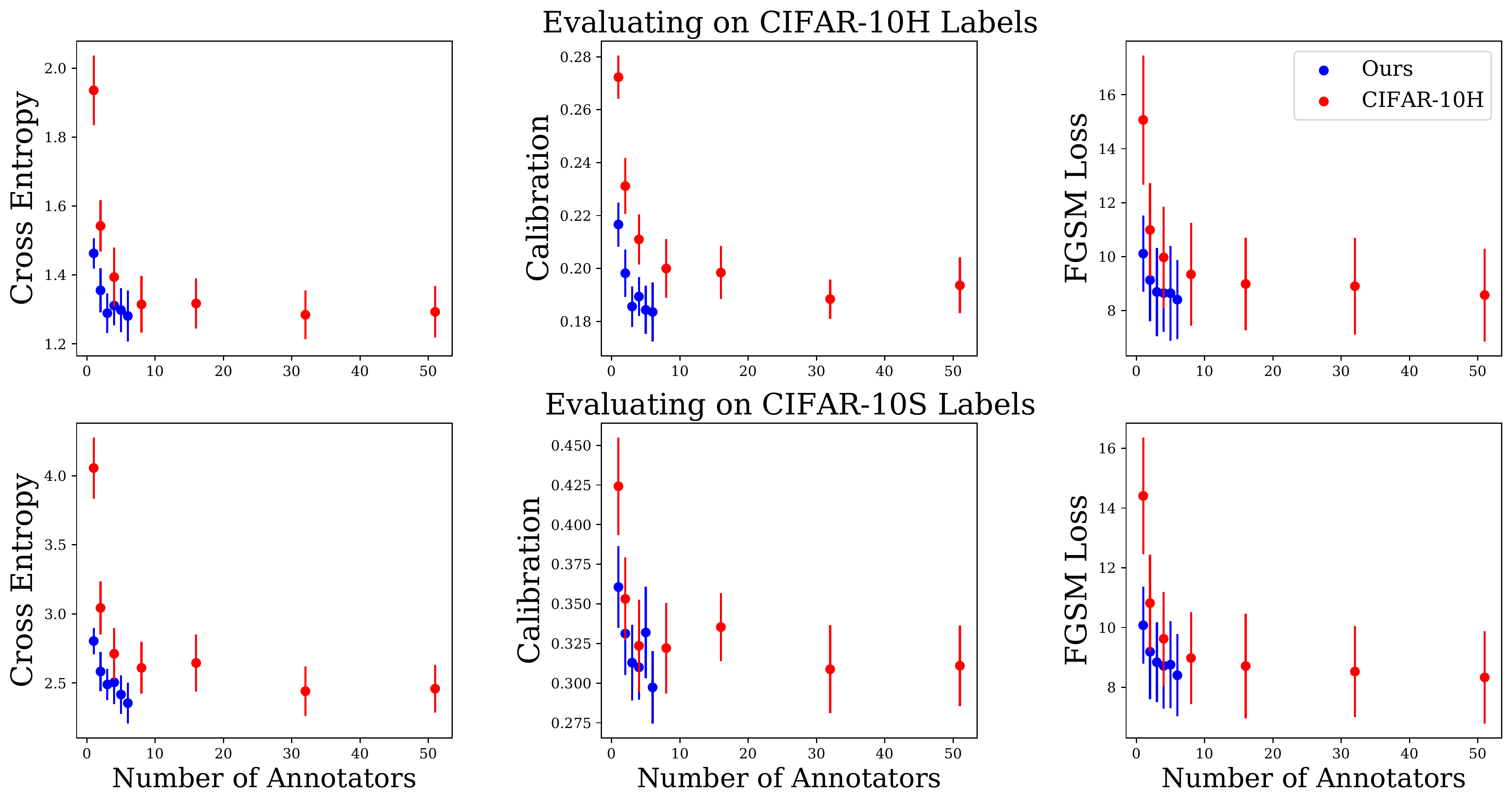}
      \caption{Comparison of learner performance based on number of annotators used to create the training labels. Red dots depict performance when aggregating $M$ \texttt{CIFAR-10H} annotators for $M \in \{1, 2, 4, 8, 16, 32, \text{all}\}$. Blue dots indicate \texttt{CIFAR-10S} T2 Clamp soft labels, constructed from varying $M \in \{1, 2, 3, 4, 5, 6\}$. Dots represent performance averaged over $15$ seeds ($5$ run per architecture type).}
       \label{fig:elicitationMCompare}
  \end{center}
  \end{figure*}

We depict a full comparison of performance when varying the number of annotators we are aggregating over in Fig. \ref{fig:elicitationMCompare}. Across most metrics and evaluation sets, our method is significantly more annotator efficient. We recognize, however, that total annotation time (accounting for the time spent per annotator) is also a practical concern. We visualize the same performance relationship with total estimated annotation time ($M * t_\text{per}$
) in the Appendix. On this cost basis, it is not clear whether our labels are advantageous.

It is possible then that we could get by with less information elicited per annotator. We explore how model performance varies as a function of the various label types that could be construed from our elicitation (see Fig \ref{fig:labelCreation}). We do find in Table \ref{tab:compare_label_varities} that while learning with all elicited information (T2,  Clamp) yields the most consistently appealing performance, we could achieve relatively good calibration and robustness in particular from less information (such as not requiring the clamp, or not eliciting probabilistic information about the second most probable label if using a Clamp) -- though more work is needed to tease apart the benefits of each elicited bit of information towards supporting a models' predictive power.

 \paragraph{\textit{Takeaways}} Eliciting and learning with individuals' soft labels -- from a few annotators -- results in a classifier that achieve better performance and robustness compared to the results of \citet{peterson2019human} who used many more annotators, particularly when \textit{de-aggegated} during training.
 %-- even so far as constructing the label from a \textit{single} annotator -- 
 This highlights that collecting categorical soft labels can be beneficial. While our method offers consistent advantages in the few-annotator regime, the benefits of eliciting per-annotator soft labels versus many annotators' hard labels is not clear when accounting for total annotation time.

\begin{table*}[htbp]
    \centering
    \begin{tabular}{lccccc} 
    \toprule
     & Label Type  & Time  & CE      &   Calibration   & FGSM Loss       \\
    \midrule
    \multirow{4}{*}{\STAB{\rotatebox[origin=c]{90}{ \texttt{10H}}}} &  T1, Unif  &  \textbf{76.8s} & 1.320$\pm$0.08  & 0.187$\pm$0.01 & 8.647$\pm$1.72     \\
 & T1, Clamp &  115.2s   & 1.299$\pm$0.06 & 0.19$\pm$0.01  & 8.385$\pm$1.55      \\
 & T2, Unif  &  153.6s & 1.287$\pm$0.10  & \textbf{0.180$\pm$0.02}  & \textbf{8.345$\pm$1.63} \\
 & T2, Clamp &  192.0s & \textbf{1.281$\pm$0.06} & 0.184$\pm$0.01 & 8.406$\pm$1.75     \\
     \midrule
     \multirow{4}{*}{\STAB{\rotatebox[origin=c]{90}{ \texttt{10S}}}} &  T1, Unif  &  \textbf{76.8s} & 2.512$\pm$0.12 & 0.317$\pm$0.04 & 9.029$\pm$1.7     \\
 & T1, Clamp & 115.2s & 2.501$\pm$0.18 & 0.306$\pm$0.03 & 8.590$\pm$1.42     \\
 & T2, Unif  &  153.6s & 2.437$\pm$0.16 & \textbf{0.293$\pm$0.04} & 8.605$\pm$1.59      \\
 & T2, Clamp &  192.0s & \textbf{2.355$\pm$0.14} & 0.297$\pm$0.03 & \textbf{8.405$\pm$1.59}   \\
    \bottomrule
    \end{tabular}

    \caption{Training models over labels constructed from subsets of the human knowledge we elicit. $M=6$ humans' labels are used to form the pool sampled. Time depicts the \textit{estimated} elicitation time (over $M=6$ annotators) for a given example.}
    \label{tab:compare_label_varities}
\end{table*}

% We depict a full comparison of performance when varying the number of annotators we are aggregating over in Fig. \ref{fig:elicitationMCompare}. Across most metrics and evaluation sets, our method is significantly more annotator efficient. We recognize, however, that total annotation time (accounting for the time spent per annotator) is also a practical concern. We visualize the same performance relationship with total estimated annotation time ($M * t_\text{per}$
% ) in Appendix Fig. \ref{fig:elicitationCostCompare}. On this cost basis, it is not clear whether our labels are advantageous. %the performance metrics of our method and that of \cite{peterson2019human} are poorer.
% %and across all label varieties in Appendix Fig. \ref{fig:elicitationCostCompareVarieties}. 

 %Future work can carefully select the images given to annotators to label: this may further demonstrate the benefits of our method. 
% We leave open the study of why our labels do not provide the same gains under PGD as FGSM.

% \paragraph{\textit{Takeaways}} Soft labels, whether constructed via many annotators' hard labels (\texttt{CIFAR-10H}) or from few annotators' soft labels (\texttt{CIFAR-10S}) improve model performance, corroborating the findings of \cite{peterson2019human, Uma_Fornaciari_Hovy_Paun_Plank_Poesio_2020}. Our labels require 8.5x fewer annotators than the labels of \texttt{CIFAR-10H} yet achieve similar performance.
% Just from eliciting and aggregating the labels of incredibly few of our annotators -- even so far as constructing the label from a \textit{single} annotator -- the resulting classifier achieves better performance and robustness. These data highlight the utility of collecting categorical soft labels -- and even partial soft labels, e.g., just for the most probable label -- is beneficial. 
% While our method offers consistent advantages in the few-annotator regime, the benefits of eliciting per-annotator soft labels versus many annotators' hard labels is not clear when accounting for total annotation time. 

% see above -- ref'd appendix fig. i think we can be transparent about that? but not take up like a half a page/section in the main text on it! we are back at 9 pgs :) 

\subsection{Simulating Annotations without Elicited Uncertainty} 
% \todo{Suppose we take just the top two labels we got, ... uniform over the top 2....} 
We have so far shown that our soft label approach yields comparable performance and even \textit{outperforms} models trained with \texttt{CIFAR-10H} labels along most metrics in the few annotator regime. This holds across several varieties of soft labels that can be formed from our labels. However, it is unclear the extent to which eliciting annotator probability estimates is beneficial. 

We address this question by training a model on labels which simulate the setting where an annotator could only select the two most probable labels (``Select Top 2 Only''). In this scenario, we assume that we do not have access to annotators' relative likelihood weightings amongst those two classes; therefore, we spread mass uniformly over the top two selected. We compare this against our labels which \textit{do} allow annotators to specify relative probabilities.
We see in Table \ref{tab:simT2_compare} that relative uncertainty information \textit{does} allow the construction of more effective labels for a learner.

\begin{table*}[htb]
\centering
    \begin{tabular}{lccccc} 
    \toprule
     & Label Type  & Time  & CE      &   Calibration   & FGSM Loss       \\
    \midrule
    \multirow{4}{*}{\STAB{\rotatebox[origin=c]{90}{ \texttt{10H}}}} &  Select Top 2 Only  &  \textbf{76.8s} & 1.300$\pm$0.06   & \textbf{0.180$\pm$0.01}  & 8.639$\pm$1.46    \\
 & Top 2 With Prob (Unif)  &  153.6s & 1.287$\pm$0.10  & \textbf{0.180$\pm$0.02}  & \textbf{8.345$\pm$1.63} \\
 & Top 2 With Prob (Clamp) &  192.0s & \textbf{1.281$\pm$0.06} & 0.184$\pm$0.01 & 8.406$\pm$1.75     \\
     \midrule
     \multirow{4}{*}{\STAB{\rotatebox[origin=c]{90}{ \texttt{10S}}}} &  Select Top 2 Only  &  \textbf{76.8s} & 2.448$\pm$0.15 & 0.313$\pm$0.02 & 8.797$\pm$1.25    \\
 & Top 2 With Prob (Unif)  &  153.6s & 2.437$\pm$0.16 & \textbf{0.293$\pm$0.04} & 8.605$\pm$1.59      \\
 & Top 2 With Prob (Clamp) &  192.0s & \textbf{2.355$\pm$0.14} & 0.297$\pm$0.03 & \textbf{8.405$\pm$1.59}   \\
    \bottomrule
    \end{tabular}
    \caption{Training with ablated annotator uncertainty over de-aggregated labels drawn from the pool of $M=6$ annotators. Here, ``Select Top 2 Only'' places equal mass on the labels selected as most probable and second most probable by each annotator.}
    \label{tab:simT2_compare}
\end{table*}

\paragraph{\textit{Takeaways}} Eliciting probability information, rather than simply selecting the most probable classes, provides useful learning signals to improve generalization and robustness. 

\section{Discussion} 

\paragraph{Fewer Annotators Needed if Eliciting Soft Labels} We demonstrate that constructing training labels from per-annotator soft labels allows practitioners to use significantly fewer labelers and still enjoy the benefits of improved model generalization, bolstered robustness, and better calibration found when aggregating many annotators' hard labels \cite{peterson2019human}. 
% The practical ramifications of this are wide. 
While online crowdsourcing platforms like Prolific~\cite{palan2018prolific} and Amazon Mechanical Turk (MTurk) enable researchers to rapidly scale experiments to many annotators, it may be challenging to recruit large numbers of annotators in domains that require expertise like medicine or criminal justice. Our elicitation approach serves to lower the barrier for efficient in-house data annotation: \textbf{\textit{when it is hard to recruit many annotators, our approach to eliciting soft labels from just a few annotators may be particularly effective.}} Such annotator efficiency could also be used to support rapid personalization. As an example, ``teachable object recognition'' is being used to enable people with visual challenges to adapt classifiers to their particular needs \cite{massiceti2021orbit}. In this application area, we may only have a single vision-impaired user per input, warranting the need for rich, single-annotator schemes such as the one we propose.

\paragraph{The Sensibility of Eliciting Annotator Probabilities} The notion of eliciting soft labels from annotators has conceptual niceties. In particular, our labeling scheme \textit{empowers} annotators to express probability judgements they have in their label assignment. In a hard label setting, annotators are required to select a single label \cite{peterson2019human}; however, an annotator has no means to express if they have any ambiguity in their label, which could occur if the image is particularly noisy or there are many similar label options. While humans have been found to have biases in their probabilistic assessments of the likelihood of phenomena~\cite{lichtenstein1977calibration, kahneman1996reality,uncertainJudgments, sharot2011optimism}, we do not think this is a sufficient reason to avoid eliciting probability judgments from annotators. As noted by \citet{uncertainJudgments} and \citet{expertElicitation}, human uncertainty can be elicited reliably as long as elicitation is rigorous. If an annotator is unsure of their decision, forcing an annotator to compress out all of this uncertainty by specifying one hard label only exacerbates, rather than solves, the challenge of capturing annotator ambiguity. 
%-- even if the exact probability they specify may not be their \textit{exact} uncertainty in the class -- 

Indeed, reasoning under uncertainty is a linchpin of human cognition \cite{lake2017building} and has been shown to be a central component of ``good'' decision-making \cite{laidlaw2021uncertain,bhatt2021uncertainty}. Careful consideration of uncertainty is of particular importance in high-stakes areas like medicine wherein diagnoses and suitable treatment plans may not be absolutely certain \cite{medUncertainty, shrager2019cancer, platts2020toleranceMedUnc, cox2021diagnostic}. Hence, annotation schemes which enable the expression of probability judgements, particularly in datasets wherein the underlying ``ground truth'' or the ``true'' label is unknown, may be sensible and desirable for improving machine safety, trustworthiness, and efficacy. Our elicitation approach takes a practical step towards this goal and offers pragmatic benefits for learning and generalization. 

% Fields wherein high-stakes decisions are frequently made like medicine are increasingly encouraging trainees to recognize and thoroughly consider when there is uncertainty in a diagnosis or treatment plan in an effort to improve medical safety \cite{medUncertainty, platts2020toleranceMedUnc, cox2021diagnostic}. 
Hence, annotation schemes which enable the expression of probability judgements, particularly in datasets wherein the underlying ``ground truth'' or the ``true'' label is unknown, may be sensible and desirable for improving machine safety, trustworthiness, and efficacy. Our elicitation approach takes a practical step towards this goal and offers pragmatic benefits for learning and generalization. 

% Fields which have traditionally de-emphasized uncertainty, such as in the medical profession, are now encouraging trainees to recognize and thoroughly consider when there is uncertainty in a diagnosis or treatment plan in an effort to improve medical safety \cite{medUncertainty, platts2020toleranceMedUnc, cox2021diagnostic}. Hence, annotation schemes which enable the expression of probability judgements, particularly in datasets wherein the underlying ``ground truth'' or the ``true'' label is unknown, may be sensible and desirable for improving machine safety, trustworthiness, and efficacy. Our elicitation approach takes a practical step towards this goal and offers pragmatic benefits for learning and generalization. 

\paragraph{Considerations of Ground Truth and Performance} 
\citet{battleday2020capturing} require their annotators to be accurate with respect to \texttt{CIFAR-10} labels. The average annotator accuracy of \texttt{CIFAR-10H} is 95\%. We do not discard annotators by accuracy alone, hence our annotator accuracy is slightly lower: annotators chose the \texttt{CIFAR-10} label as the top label approximately 84\% of the time and include this label in their top two choices 92\% of the time. We hope our dataset may be helpful for researchers studying learning from semi-noisy annotators. 

While many ML tasks assume a `true' label to calculate metrics like accuracy, there are settings where it is not possible or sensible to aim for a single true label agreed upon by all annotators. It is unclear what should count as ``ground truth'' in how we evaluate models. Our soft labels permit multiple categories to be simultaneously considered without explicit consensus, and therefore offer an alternative form of evaluation potentially well-suited for such settings. 

% have potential to serve as interesting evaluation sets for future work.

% In future work, we hope to further leverage probabilistic judgements to reconsider notions of `truth.' 

% To be conservative, and for consistency with \cite{peterson2019human}, \textbf{we report results for our methods considering the labels of \texttt{CIFAR-10H} as `truth'}\todo{revise section now that we eval on \texttt{10S}}. It is promising that we perform well in this setting. We expect our methods to perform better if a held out set of our own \texttt{CIFAR-10S} labels is instead assumed to be `true', as might be considered appropriate. % or other labelings to define good performance. 
% %Unlike existing representations of labels and ground truth, 
% Our soft labels permit multiple categories to be simultaneously considered without explicit consensus. In future work, we hope to further leverage probabilistic judgements to reconsider notions of `truth'. %what the ``true'' label of an image means.

\paragraph{Limitations} While our labeling methods yield conceptual benefits, and potential advantages in annotator efficiency and improved performance, we recognize that currently our approach is significantly slower to collect per annotator than hard labeling. Moreover, we have only considered a single annotation framework, and as we have seen, model performance is sensitive to the structure of the provided labels. It is therefore possible that alternative elicitation paradigms, such as those wherein annotators select more than just their inferred top 2 most probable categories or even provide soft labels graphically through a constructed histogram \citep{goldstein2014lay}, could yield different labels and hence impact model training. And while we did not find significant intra-annotator variability on repeat trials,\footnote{Only approximately 7\% of people changed their most probable label between repeated instances. Of people who did not change their most probable category selection, the average change in prob was 6\%.} it is important to consider whether annotators' internal label distributions are stable over time \citep{murray2015posterior}. We also note that even though our elicitation framework enables us to learn effectively from fewer annotators, too few annotators could yield various biases, especially as ``50\%'' to one annotator may not mean the same to another, and the inherent task difficulty could impact whether more annotators \textit{ought} to be recruited -- even if effective learning in the current study can be achieved from few annotators. 

Indeed, we are cognizant that our findings are within a particular domain, image classification,  over a particular dataset, the test set of \texttt{CIFAR-10} (i.e., we collect annotations over $10\%$ of \texttt{CIFAR-10H}). Here, we have a managable number of categories to elicit information over; it is hard to imagine an annotator selecting \textit{all} impossible categories if there are 100s or 1000s as may be the case in other ML datasets \citep{fei2009imagenet}. More work is needed to verify if our results generalize to other settings, and extend beyond the crowdsourcing space to real-world domain experts. All participants considered here are based in the United States and speak English as their first language. As discussed in \cite{prabhakaran2021releasing, diaz2022crowdworksheets}, recruiting a diverse group of annotators from different backgrounds and releasing disaggregated annotator responses is valuable to ensure a broad spectrum of human experiences and world views are captured in datasets.

% \todo{add notes about task difficulty, annotator noise, and alternative interfaces (poss in next section)}

\paragraph{Using and Extending our Interface} To design more time-efficient elicitation and scale our soft label elicitation interface to other domains, we make our elicitation interface publicly available.\footnote{Our code can be found at: \url{https://github.com/cambridge-mlg/cifar-10s/}.} To apply our set-up to a new problem, all one needs are: 1) a folder of the images one wishes to present to the annotator, 2) a set of labels that the annotator is allowed to select, and 3) an allocation of images to batches (e.g., a \texttt{.json} file). We hope the ease with which our interface can be adapted to new domains will lower the barrier of entry for others to run their own soft label data collection.  

\paragraph{Additional Extensions} Future work can consider how the elicitation of soft labels via annotator probabilities may change over a broader set of datasets and domains, and may alter if eliciting across a wider spectrum of annotator backgrounds, where some may be assumed to have access to ``privileged information'' based on their experiences which are worthy to model \cite{Sharmanska_2016_CVPR}. And in light of the time costs of our elicitation, we see promise in developing active methods to identify which images may benefit most from being queried via our rich elicitation scheme. %, and which may suffice to be tagged with hard labels. 
We encourage researchers and designers to create more time-efficient schemes to elicit rich annotator probability or uncertainty measures towards constructing good soft labels. 
% We see potential in our elicitation schemes being used to garner more fine-grained human confidence information to support probabilistically-grounded teaming approaches in an annotator-efficient manner~ \cite{steyvers2022bayesian}\textcolor{red}{??}.

We have only considered simple naive averaging to aggregate annotations; future work could draw on the expansive literature concerning aggregation \cite{dawid1979maximum, smyth1994knowledgeUncImgs,vote, whitehill2009whoseVote, ho2016eliciting, Sharmanska_2016_CVPR, augustin2017bayesian, zhang2018multi, wang2021forecast, wei2022aggregate, collier2022transfer} to develop better schemes which may take into account differential annotator characteristics such as trustworthiness and expertise. 
%Our redistribution strategies  only considered uniform smoothing over all labels deemed possible; however, more sophisticated smoothing practices which abide by class relationships could be employed. 
We encourage researchers to evaluate the efficacy of our constructed soft labels in other learning paradigms, such as weakly supervised learning \cite{arazo2019unsupervised, wei2022aggregate}, online learning~\cite{chen2022perspectives}, or curriculum learning~\cite{liu2017iterative}, and as human-grounded priors in Bayesian neural network settings \cite{fortuin2022priors}.
% Similarly, future work can consider how the elicited labels impact model adaptability in a setting where expert feedback is received online~\cite{chen2022perspectives}. 

% other redistribution methods?

% Indeed, such annotator efficiency can support personalization. 

% , future work ought to consider scaling our elicitation experiment to real-world domains, for instance, in the medicalfull scale? towards more ``real-world'' datasets, other elicitation schemes, incorporating human uncertainty not just to improve machine classification performance, but using these fine-grained representations of human confidence for teaming like steyvers. could consider expert weightings (cite work)? other bounds? other redistribution methods? synthesizing examples? 

% While humans have been found to be miscalibrated in their subjective reports of uncertainty \todo{[cite]}, we do not think this is a sufficient reason to avoid eliciting uncertain judgments from annotators. Indeed, if . Moreover, the practice of 

\section{Conclusion}

In this work, we have shown the benefits of eliciting and aggregating per-annotator soft labels over aggregating hard labels on the \texttt{CIFAR-10} dataset~\cite{peterson2019human}.  The benefits we observe include: % Learning over elicited soft labels enables us to
improved model generalization, calibration, and robustness from fewer total annotators; and %. We find that the information contained in our soft labels provides improved 
richness in the learning signal which enables improved calibration. 
We release the code for our elicitation interface and our collected soft label dataset as \texttt{CIFAR-10S}. 
We hope that our work encourages others to explore the benefits of eliciting soft labels from annotators.

%and that our experiments motivate machine learning practitioners to seek out such datasets to improve model training, robustness, and general machine trustworthiness.  

% mention hierarchies or leave separate? 

\section*{Acknowledgments} 
We thank (alphabetically) Krishnamurthy (Dj) Dvijotham, Carl Henrik Ek, Weiyang Liu, Bradley Love, Vihari Piratla, Jeff Shrager, Richard E. Turner, Joshua Tenenbaum, Marty Tenenbaum, and Miri Zilka for helpful discussions. We also thank Ruairidh Battleday and Joshua Peterson for helpful clarifications on the \texttt{CIFAR-10H} elicitation, as well as Alexandra Uma for sharing the code for their paper \cite{Uma_Fornaciari_Hovy_Paun_Plank_Poesio_2020}. We also thank our reviewers for very helpful feedback on our manuscript. 

KMC is supported by a Marshall Scholarship. UB acknowledges support from DeepMind and the Leverhulme Trust via the Leverhulme Centre for the Future of Intelligence (CFI), and from the Mozilla Foundation. AW acknowledges support from a Turing AI Fellowship under grant EP/V025279/1, The Alan Turing Institute, and the Leverhulme Trust via CFI.

\bibliography{main} 

\clearpage
\section*{Appendix} 
% Our appendix solely consists of support figures and their respective captions, all of which are referenced in text.

\begin{figure*}[htb]
  \begin{center}
  \includegraphics[width=0.9\linewidth]{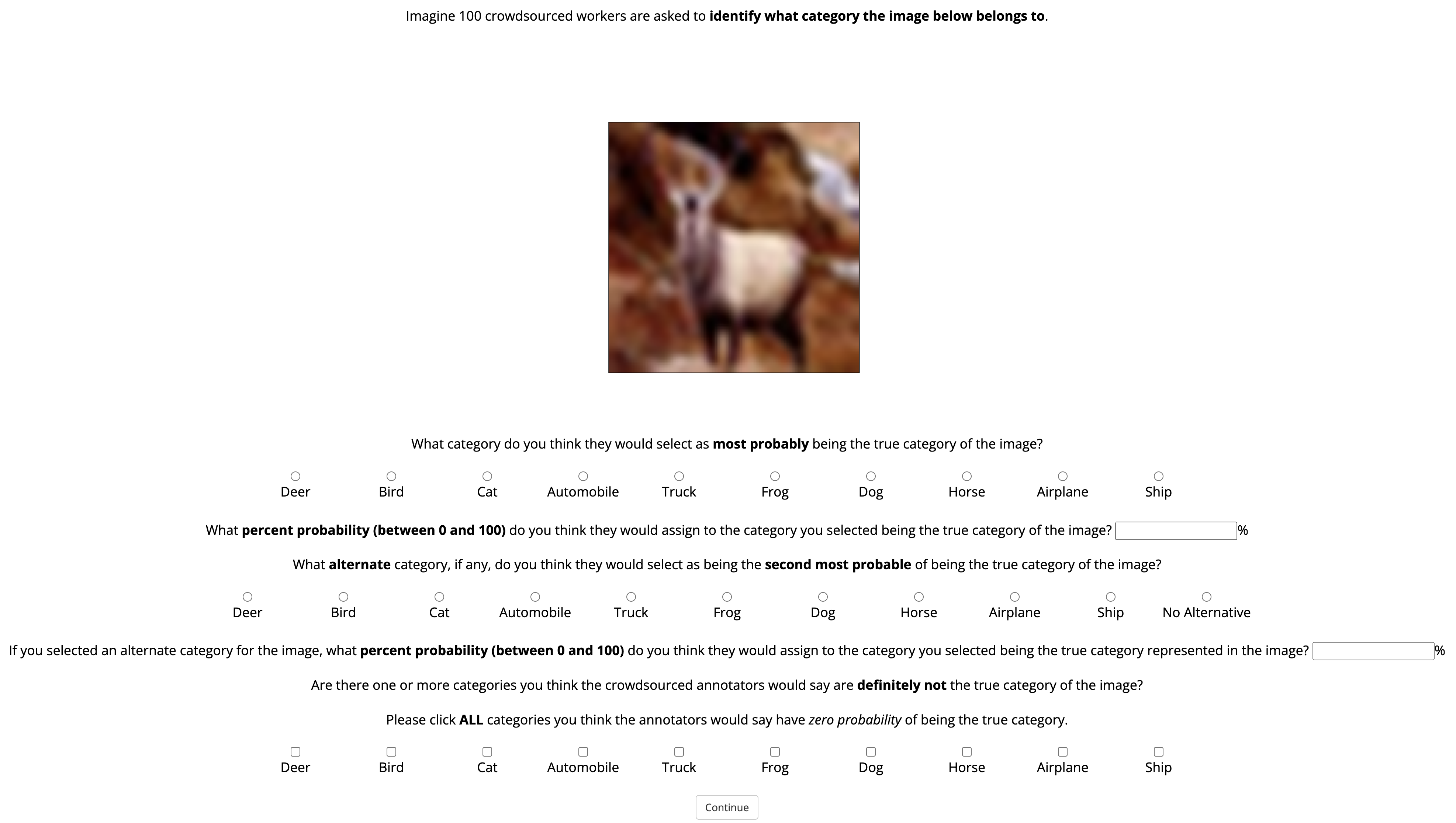}
      \caption{Depiction of our soft label elicitation interface.}
       \label{fig:elicitationInterface}
  \end{center}
  \end{figure*}

\begin{figure*}[b!]
  \begin{center}
  \includegraphics[width=0.8\linewidth]{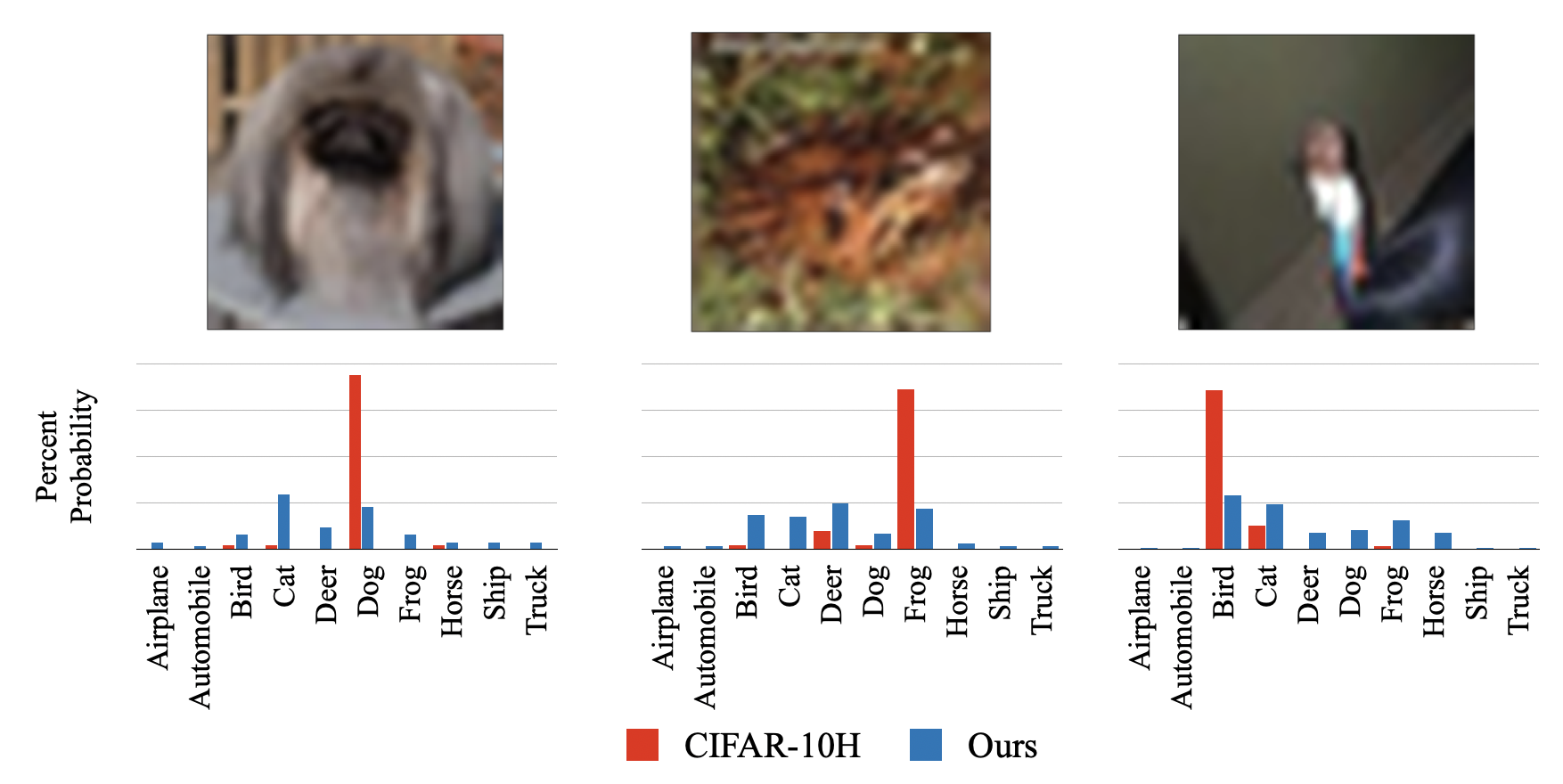}
      \caption{Top three highest Wasserstein distance examples between our \texttt{CIFAR-10S} labels (blue) and \texttt{CIFAR-10H} (red). The hard labels in \texttt{CIFAR-10} are: dog, frog, and bird.}
  \label{fig:compareLabelsHighDist}
  \end{center}
  \end{figure*}

\begin{figure*}[htb]
  \begin{center}
  \includegraphics[width=0.8\linewidth]{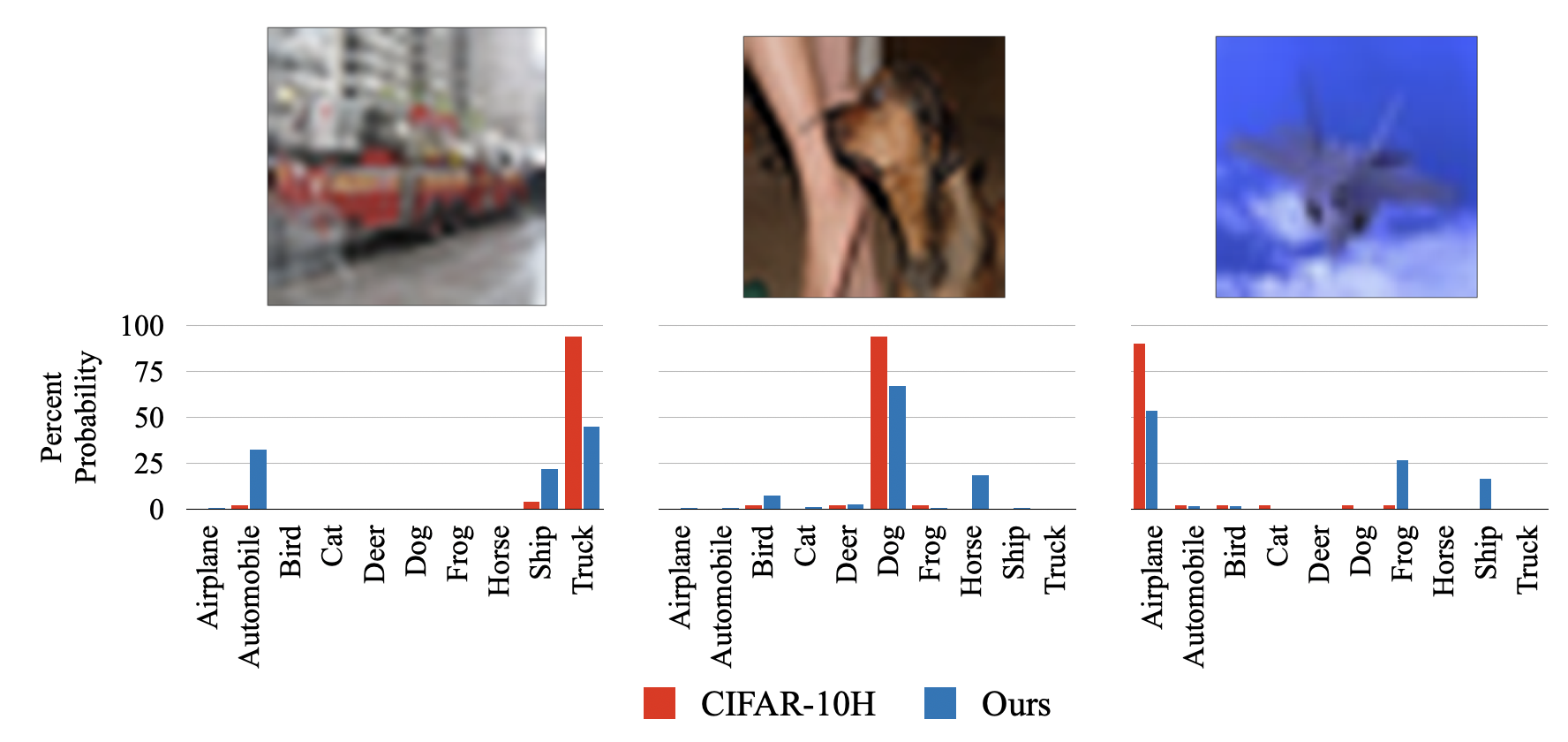}
      \caption{Additional examples demonstrating the ``softening'' of \texttt{CIFAR-10H} labels with our elicitation. The hard labels in \texttt{CIFAR-10} are: truck, dog, and airplane.}
  \label{fig:compareLabelsSoftening}
  \end{center}
  \end{figure*}

\begin{figure*}[htb]
  \begin{center}
  \includegraphics[width=1.0\linewidth]{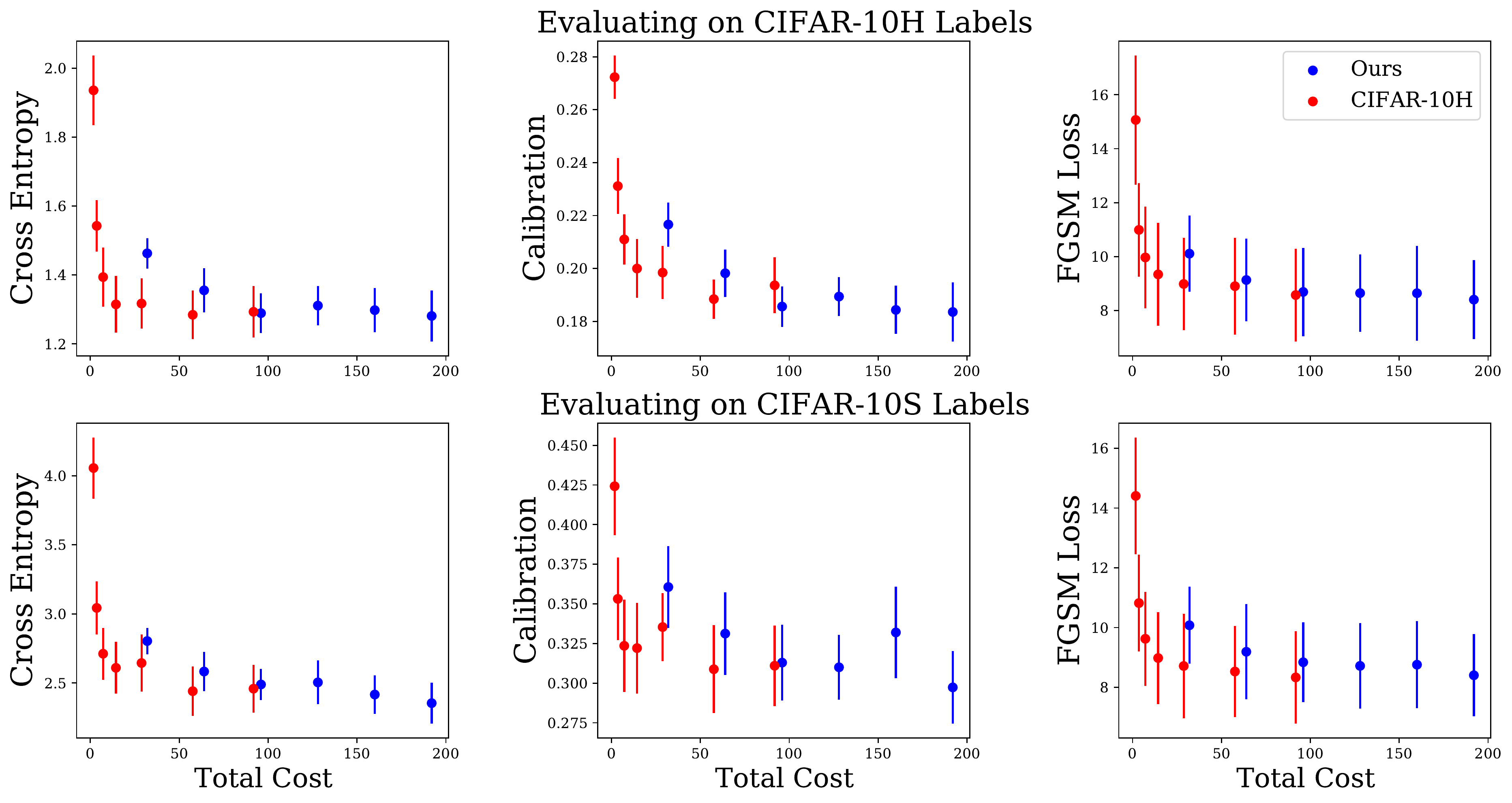}
      \caption{Comparison of learner performance as a factor of \textit{estimated} total cost of elicitation ($M * t_\text{per}$). Red dots depict performance when aggregating $M$ \texttt{CIFAR-10H} annotators for $M \in \{1, 2, 4, 8, 16, 32, 51\}$. Blue dots indicate \texttt{CIFAR-10S} T2 Clamp soft labels, constructed from varying $M \in \{1, 2, 3, 4, 5, 6\}$. Dots represent performance averaged over $15$ seeds ($5$ run per architecture type).}
        \label{fig:elicitationCostCompare}
  \end{center}
  \end{figure*}

This Appendix includes: 

\begin{enumerate}
    \item Our elicitation interface shown to recruited participants in Fig. \ref{fig:elicitationInterface}.
    \item Additional elicited label distributions in Figs. \ref{fig:compareLabelsHighDist} and \ref{fig:compareLabelsSoftening}.
    \item A comparison against classical label smoothing in Table \ref{tab:label_smooth_compare}, with a discussion of the results in the following section. 
    \item Performance as a factor of estimated total annotation time in Fig. \ref{fig:elicitationCostCompare}.
\end{enumerate}

\begin{table*}[htbp]
    \centering
    \begin{tabular}{ccccc} 
    \toprule
     & Label Type       & CE             & Calibration    & FGSM Loss       \\
    \midrule
    \multirow{2}{*}{\STAB{\rotatebox[origin=c]{90}{ \texttt{10H}}}} & Label Smoothing  & 1.368$\pm$0.19 & \textbf{0.175$\pm$0.05} & \textbf{6.965$\pm$1.7} \\
    & CIFAR-10H       & 1.293$\pm$0.08 & 0.194$\pm$0.01 & 8.577$\pm$1.91 \\
  & Ours (T2, Clamp) & \textbf{1.281$\pm$0.06} & 0.184$\pm$0.01 & 8.406$\pm$1.75 \\
     \midrule
     \multirow{2}{*}{\STAB{\rotatebox[origin=c]{90}{ \texttt{10S}}}} & Label Smoothing  & 2.674$\pm$0.33 & 0.299$\pm$0.05 & 9.375$\pm$3.4 \\
     & CIFAR-10H       & 2.459$\pm$0.21 & 0.311$\pm$0.02 & \textbf{8.334$\pm$1.75} \\
  & Ours (T2, Clamp) & \textbf{2.355$\pm$0.14} & \textbf{0.297$\pm$0.03} & 8.405$\pm$1.59 \\
    \bottomrule
    \end{tabular}

    \caption{Comparing de-aggregated human-derived soft labels against label smoothing.}
    \label{tab:label_smooth_compare}
\end{table*}

  \subsection*{Relationship to Classical Label Smoothing} 
When considering the value of eliciting and incorporating additional human knowledge in ML systems, one may ask why not just use traditional label smoothing (LS)? 

$$y^\text{LS}_n = y^{\text{hard}}_n * (1-\beta) + \beta * y^\text{smoother}$$

\noindent $y^\text{smoother}$ is a uniform of length $K$ (each ``probability'' $= \frac{1}{K}$ (applied to all N examples), $y^{\text{hard}}_n$ is the conventional one-hot label, and $\alpha$ is the smoothing factor $\in [0,1]$. 

We train the three models considered in Section 5.1 over labels on which LS ($\beta=0.05$) was applied. We tune $\beta$ ($\in \{0, 0.0001, 0.001, 0.01, 0.05, 0.1, 0.2, 0.3, 0.4\})$) with the same validation method discussed. We find in Table \ref{tab:label_smooth_compare} that LS does outperform both \texttt{CIFAR-10H} and \texttt{CIFAR-10S} in terms of calibration and roubstness over \texttt{CIFAR-10H}; however, this does not hold when evaluated on \texttt{CIFAR-10S}. While these results do shed light on a nuance of human knowledge elicitation (alternative, automated ML approaches are powerful already); we argue that human knowledge still contributes valuable insights. Importantly, LS does not capture \textit{meaningful softness}, i.e., an image most likely to be a deer has equal probability of alternatively being a dog as a ship. This lack of human-sensible alternatives may prohibit effective generalization to our richer, harder \texttt{CIFAR-10S} evaluation set; a result of which has been shown on other datasets \cite{zhang2021delvingLS}. Furthermore, the blanket mass spread over all $K-1$ alternatives also has been found to lead to oversimplified clusters \cite{muller2019labelSmoothHelp}. It is worth investigating the kinds of latent spaces that result from training over our human-derived soft labels instead.

    % \subsection*{Elicitation Interface}

%   \subsection*{Additional Recovered Human Label Distributions}

\end{document}